\documentclass[a4paper,fleqn]{cas-dc}

\usepackage[numbers]{natbib}

\makeatletter
\renewcommand{\printemails}{}
\renewcommand{\printurls}{}
\renewcommand{\printfacebook}{}
\renewcommand{\printtwitter}{}
\makeatother

\def\tsc#1{\csdef{#1}{\textsc{\lowercase{#1}}\xspace}}
\tsc{WGM}
\tsc{QE}
\tsc{EP}
\tsc{PMS}
\tsc{BEC}
\tsc{DE}

\begin{document}
\let\WriteBookmarks\relax
\def\floatpagepagefraction{1}
\def\textpagefraction{.001}
\shorttitle{3D Skeleton-Based Action Recognition: A Review}

\author[1]{Mengyuan Liu}
\affiliation[1]{organization={State Key Laboratory of General Artificial Intelligence, Peking University, Shenzhen Graduate School},
            city={Shenzhen},
            country={China}}
            
\author[1]{Hong Liu}    
\author[1]{Qianshuo Hu} 
\author[2]{Bin Ren} 
\author[3]{Junsong Yuan} 
\author[1]{Jiaying Lin} 
\author[4]{Jiajun Wen}

\affiliation[2]{organization={Department of Information Engineering and Computer Science (DISI), University of Trento},
            city={Trento},
            country={Italy}}

\affiliation[3]{organization={University at Buffalo, The State University of New York},
            city={Buffalo, NY},
            country={USA}}

\affiliation[4]{organization={Tsinghua Shenzhen International Graduate School, Tsinghua University},
            city={Shenzhen},
            country={China}}

\title [mode = title]{3D Skeleton-Based Action Recognition: A Review}

\maketitle

\begin{abstract}
With the inherent advantages of skeleton representation, 3D skeleton-based action recognition has become a prominent topic in the field of computer vision. However, previous reviews have predominantly adopted a model-oriented perspective, often neglecting the fundamental steps involved in skeleton-based action recognition. This oversight tends to ignore key components of skeleton-based action recognition beyond model design and has hindered deeper, more intrinsic understanding of the task.
To bridge this gap, our review aims to address these limitations by presenting a comprehensive, task-oriented framework for understanding skeleton-based action recognition. We begin by decomposing the task into a series of sub-tasks, placing particular emphasis on preprocessing steps such as modality derivation and data augmentation. The subsequent discussion delves into critical sub-tasks, including feature extraction and spatio-temporal modeling techniques. Beyond foundational action recognition networks, recently advanced frameworks such as hybrid architectures, Mamba models, large language models (LLMs), and generative models have also been highlighted. Finally, a comprehensive overview of public 3D skeleton datasets is presented, accompanied by an analysis of state-of-the-art algorithms evaluated on these benchmarks.
By integrating task-oriented discussions, comprehensive examinations of sub-tasks, and an emphasis on the latest advancements, our review provides a fundamental and accessible structured roadmap for understanding and advancing the field of 3D skeleton-based action recognition.
\end{abstract}






\section{Introduction}

Action recognition is a critical component and one of the most actively researched topics in computer vision, having been explored for several decades. Its capability to identify human actions holds immense promise across diverse applications, including intelligent surveillance, human-computer interaction, virtual reality, robotics, and more \cite{khan2024human, gammulle2023continuous, li2023action, dallel2020inhard}. 
Significant advancements have been achieved using various data representations \cite{lin2019tsm, liu2020grouped, feichtenhofer2019slowfast, tran2018closer, xu2017lie, baek2016kinematic, poppe2010survey, simonyan2014two, feichtenhofer2016convolutional, liu2023explore}, such as RGB image sequences, depth images, and optical flows. However, these modalities often involve higher computational costs, reduced robustness in complex environments, and increased sensitivity to variations like body scale, viewpoint changes, and motion speeds \cite{johansson1973visual}. Moreover, they can introduce privacy concerns in practical applications.
In contrast, skeleton data offers a topological and abstract representation of the human body through joints and bones. This modality is not only computationally efficient but also less affected by environmental complexities. Recent advancements in sensors, such as Microsoft Kinect \cite{zhang2012microsoft}, and state-of-the-art human pose estimation algorithms \cite{chu2017multi, yang2016end, cao2019openpose, song2021human}, have enabled the accurate capture of 3D skeleton data \cite{si2019attention}, making it a promising alternative for action recognition.


\begin{figure*}[t]
	\centering
	\includegraphics[width=1\linewidth]{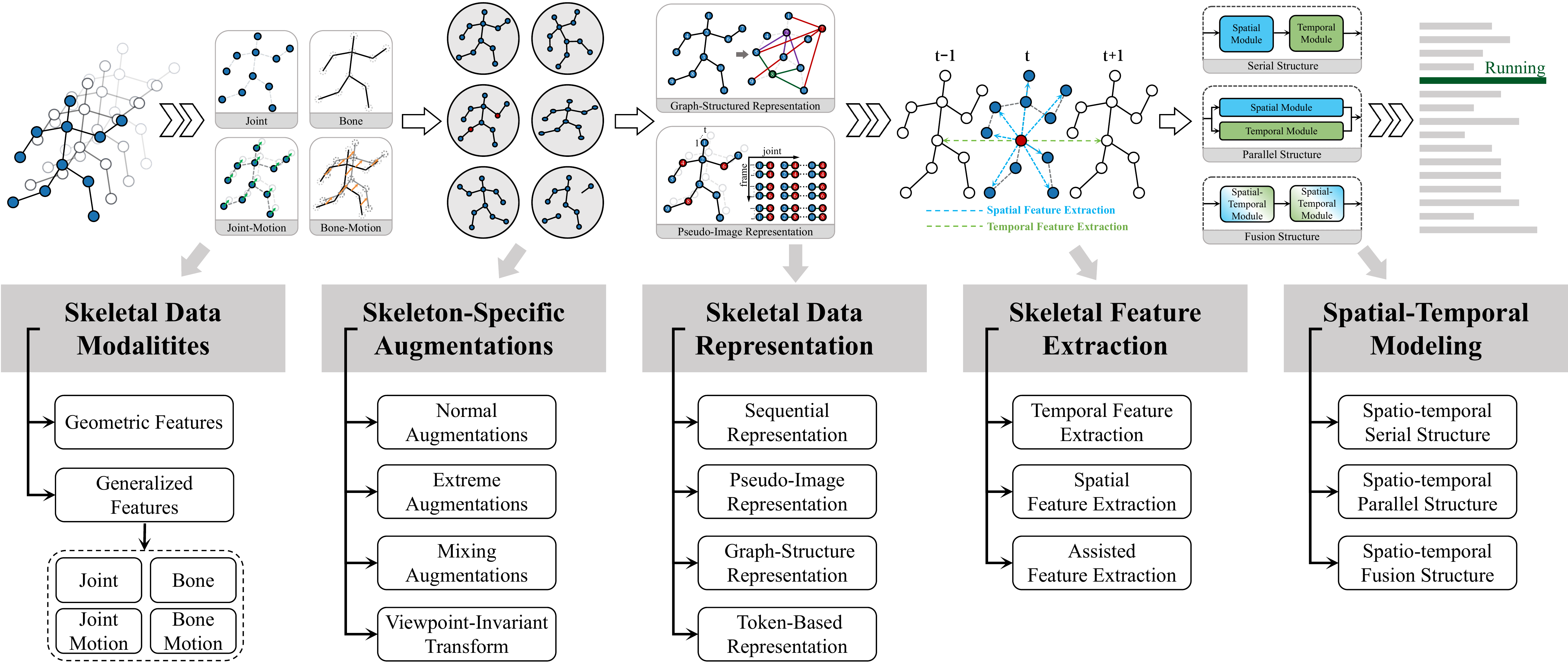   }
	\caption{A comprehensive workflow for 3D skeleton-based action recognition using deep learning}
	\label{fig:intro}
\end{figure*}

\begin{figure*}[t]
	\centering
	\includegraphics[width=1\linewidth]{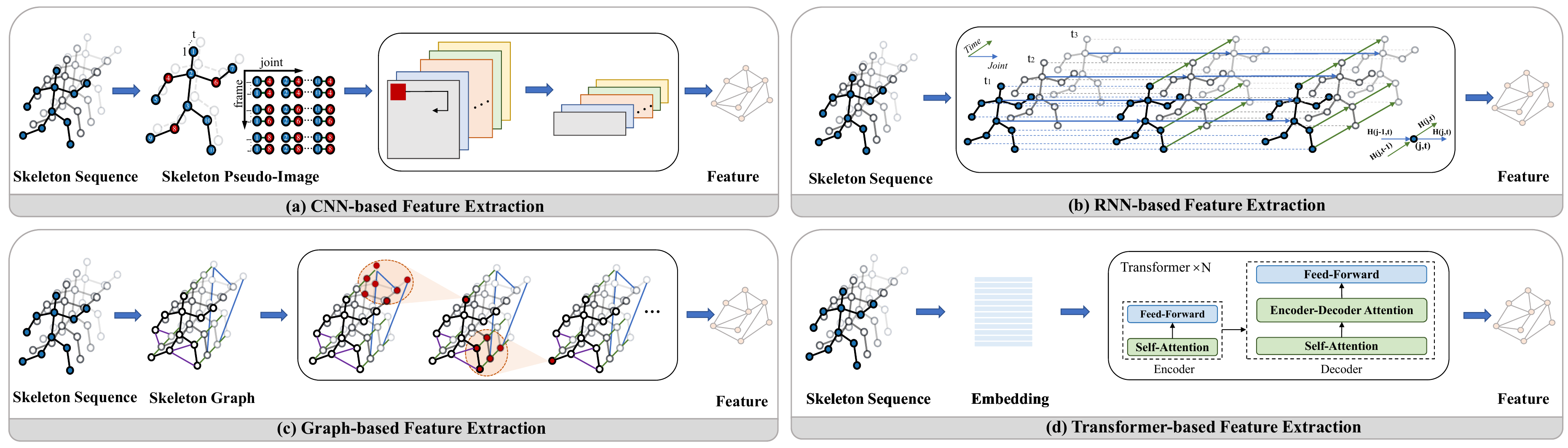}
	\caption{Main feature extraction methods}
	\label{fig:Feature}
\end{figure*}

Skeleton sequences exhibit three notable characteristics: (1) Spatial human pose: Strong correlations between joint nodes and their neighbors reveal rich structural information within each frame. (2) Temporal human action: Inter-frame correlations capture significant temporal dynamics. (3) Spatio-temporal co-occurrence: Joint and bone interactions link spatial and temporal domains. These features have garnered significant attention in human action recognition and detection, driving the growing adoption of skeleton data in the field.

Human action recognition based on skeleton sequences is primarily a temporal pattern recognition problem, with traditional methods often focusing on extracting motion patterns from skeleton data. This has led to extensive research on handcrafted features \cite{hu2015jointly,vemulapalli2014human,hussein2013human}, commonly involving relative 3D rotations and translations between joints or body parts \cite{liu2017enhanced,Vemulapalli2016Rolling}. However, studies have shown that handcrafted features typically perform well only on specific datasets \cite{wang2019comparative}, highlighting the issue that features designed for one dataset may not transfer effectively to others, especially in wild environments. This limits the generalizability and broader applicability of action recognition algorithms.

\begin{figure*}[t]
	\centering
	\includegraphics[width=1\linewidth]{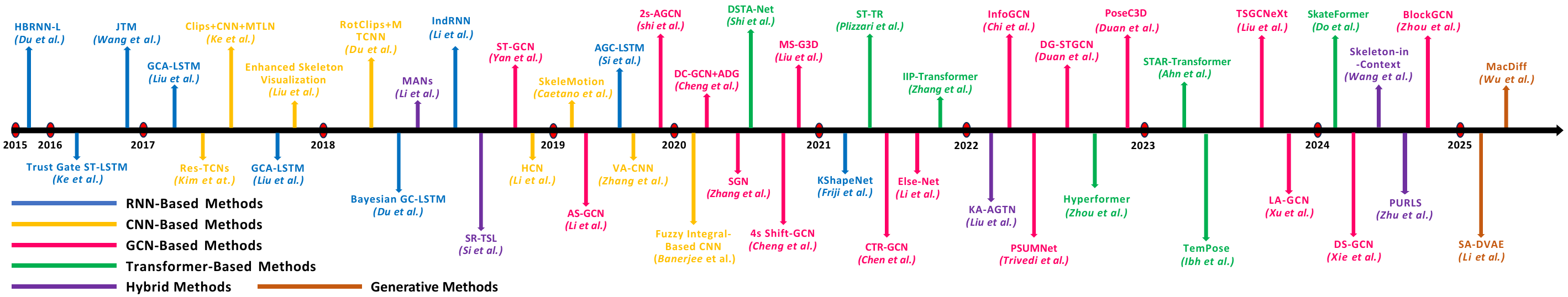}
	\caption{Chronological overview of the most relevant deep learning-based action recognition methods.}
	\label{fig:fig15}
\end{figure*}

With the rapid advancements in deep learning, methods such as RNNs, CNNs, GCNs, Transformers, hybrid methods and generative methods, have been applied to skeleton-based action recognition, as Fig.\ref{fig:fig15}. RNN-based models \cite{du2015hbrnn,liu2016tree,song2017end,wang2017modeling}, like LSTMs and GRUs, are strong in temporal modeling, treating skeleton sequences as time-series data. CNNs \cite{ke2017new,kim2017interpretable,li2018co} excel at capturing spatial relationships among joints by encoding skeleton sequences as pseudo-images. However, both RNNs and CNNs fail to model the intrinsic relationships among joints. To address this, GCNs \cite{yan2018stgcn,si2018srtsl,shi2019skeleton} represent skeleton data as graphs, where joints are vertices and bones are edges, offering better representation of structural interactions. 
Transformers \cite{shi2020decoupled,cho2020self,zhang2021stst,xin2023transformer}, with their self-attention mechanism, have gained traction for modeling joint relationships and improving skeleton data analysis by replacing handcrafted adjacency matrices.
More recently, hybrid methods \cite{liu2022graph, yang2023stream, xin2023mixformer, duan2023skeletr, liu2024HDBN} seek to combine the strengths of two or more backbone architectures, such as RNNs, CNNs, GCNs, and Transformers, to more comprehensively model both spatial and temporal aspects of skeleton data. 
Generative methods \cite{yan2023skeletonmae, wu2025macdiff, lin2025idempotent, li2025sa} have been adapted to pre‐train and regularize skeleton representations by reconstructing or predicting masked joints, thereby capturing intrinsic spatio‐temporal patterns in an unsupervised manner.     

Skeleton-based action recognition using deep learning methods has been extensively summarized and discussed in several surveys \cite{ren2024survey, sun2022human, xin2023transformer, shin2024comprehensive}. 
Sun et al. \cite{sun2022human} and Shin et al. \cite{shin2024comprehensive} provide detailed discussions on the Action Recognition task under diverse data modalities, but they only briefly cover 3D skeleton-based methods without in-depth analysis. Xin et al. \cite{xin2023transformer} specifically focus on 3D skeleton-based methods, but their discussion is limited to Transformer-based approaches. Ren et al. \cite{ren2024survey} offer a comprehensive overview, covering methods ranging from RNNs to Transformers. These surveys typically categorize research efforts based on model architectures, such as RNNs, CNNs, GNNs, and Transformers. 

However, existing reviews still exhibit several shortcomings. On one hand, although categorizing methods by architecture effectively traces the field’s evolution and highlights focal points over time, recent advances have given rise to more flexible hybrid designs \cite{liu2022graph, yang2023stream, xin2023mixformer, duan2023skeletr, liu2024HDBN} that exploit the unique properties of skeletal action sequences and transcend the constraints of traditional models. Moreover, these surveys have yet to cover emerging approaches such as those based on large language model prompting or Mamba architectures. On the other hand, as illustrated in Fig. \ref{fig:intro}, each stage, from skeleton data processing through to spatio-temporal feature modeling, critically influences final performance. Yet existing reviews tend to concentrate almost exclusively on the modeling phase, without offering a comprehensive overview of the data preprocessing steps, which we believe deserves equal attention.
These observations lead to three key insights: (1) classification by model architecture alone is insufficient; instead, a systematic review of the task’s essential stages can provide a more holistic perspective and help readers delve into the core modeling challenges; (2) skeleton data preprocessing has a profound impact on subsequent spatio-temporal modeling and therefore warrants detailed examination; and (3) given the limited number of current surveys, our review must encompass the very latest developments, including Mamba architectures and large language model–based prompting.

Therefore, this paper shifts the focus from model architecture-based classification to the more fundamental challenges of skeleton-based action recognition: how to transform unstructured data into structured representations and effectively model the spatio-temporal features of skeletal sequences. Specifically, we organize our discussion along the hierarchical tasks involved in deep learning-based skeletal action sequence modeling: \textbf{derived modalities}, \textbf{data augmentation}, \textbf{data representation}, \textbf{temporal/spatial modeling}, and \textbf{spatio-temporal co-modeling}. This paradigm not only outlines the entire pipeline of this task but also provides a comprehensive analysis of the core modeling challenges in this domain. It aims to offer readers a deeper understanding of the characteristics of skeletal data and the inherent research difficulties of the task.

In summary, our paper contains four contributions:
\begin{itemize}
	\item \textbf{A more essential framework: }This review adopts a task-oriented framework, departing from the traditional model architecture-centered discussions. As shown in Fig. \ref{fig:intro}, we organize the discussion based on the key stages of the task, with a particular emphasis on data representation, skeleton modeling, and spatio-temporal co-modeling. This unique viewpoint helps readers gain a deeper understanding of the intrinsic nature of the task.
	\item \textbf{A more comprehensive framework: }We believe that the preprocessing of skeleton data plays a crucial role in subsequent spatio-temporal modeling. Therefore, we focus on and analyze tasks such as skeleton-based derived modalities and skeleton-specific augmentations in the data preprocessing phase. This allows us to refine the framework of existing surveys, which typically focus solely on spatio-temporal modeling methods, and provides readers with a more comprehensive and holistic view of the entire process.
	\item \textbf{Core Issue Refinement: }
    By breaking down the task into multiple progressively detailed subtasks, this review provides a clear research roadmap. As shown in Fig. \ref{fig:Feature}, this review makes a more essential and detailed discussion of key processes such as feature extraction and modeling, which helps identify common challenges in the entire task and encourages researchers to explore model design and optimization from the perspective of data understanding.
	\item \textbf{Exploring Cutting-Edge Methods: }This review focuses on the latest and most advanced deep learning methods for skeleton-based human action recognition, with particular attention to emerging hybrid architecture models, Mamba architecture models, and large language models (LLMs). A systematic analysis and evaluation of current state-of-the-art (SOTA) models are presented, exploring their performance and applicability to specific tasks. This in-depth discussion also provides insights into the future directions of research in this field.
\end{itemize}

The structure of the remaining sections is organized as follows. Section 2 analyzes the theoretical modeling of the overall process. Sections 3 to 7 are structured in the sequential stages of the skeleton-based action recognition task. Section 3 explores the generation and application of multimodal skeleton data. Section 4 introduces skeleton-specific data augmentation methods. Section 5 discusses techniques for converting skeleton data into representations suitable for model input. Section 6 provides a detailed account of feature extraction methods. Section 7 focuses on spatio-temporal modeling techniques. Section 8 lists commonly used datasets and the performance of state-of-the-art models on these datasets. Finally, Section 9 discusses the potential future developments in the field of skeleton-based action recognition.

\section{Overall Process Theoretical Modeling}
In this section, we will model the overall skeleton action recognition process in combination with Fig.\ref{fig:intro}, aiming to provide readers with a clear and intuitive framework through mathematical representation and step-by-step analysis. Starting from the input skeleton data modality, we will discuss key links such as data enhancement, feature representation, feature extraction, and spatio-temporal modeling. Then, we will reveal the overall process and internal mechanism of skeleton action recognition based on deep learning.

\subsection{Skeletal Data Modalities}

In skeleton-based action recognition, the input data is typically represented as a skeleton sequence, consisting of joint coordinates evolving over time, as shown in Eq.\ref{eq:xt}.
\begin{equation} \label{eq:xt}
    \mathcal{X} = \{X_t \mid t = 1, 2, \dots, T\}, \quad X_t \in \mathbb{R}^{J \times D}
\end{equation}
where $T$ is the total number of time steps, $J$ represents the number of joints, and D denotes the dimensionality of each joint (e.g., 2D or 3D coordinates). The features can be categorized into two main types: geometric features and generalized features. 
The generalized features are categorized into four distinct modalities: \textbf{Joint ($X_t^{\text{joint}}$), Bone ($X_t^{\text{bone}}$), Joint Motion ($X_t^{\text{jm}}$)}, and B\textbf{one Motion ($X_t^{\text{bm}}$)}. Each modality represents a specific characteristic of skeletal data. The Joint modality captures the positional information of key body joints, while the Bone modality encodes the relative relationships between connected joints. Joint Motion and Bone Motion modalities describe the dynamic changes in joints and bones over time. The Bone modality is defined by Eq. \ref{eq: bone}, the Joint Motion modality by Eq. \ref{eq: jm}, and the Bone Motion modality by Eq. \ref{eq: bm}, where $X_t^{\text{joint}_1}$ and $X_t^{\text{joint}_2}$ denote the two joints at the ends of a bone, and $X_{t-1}$ indicates the position of a joint or bone at the previous time step.

\begin{equation} \label{eq: bone}
    X_t^{\text{bone}} = X_t^{\text{joint}_1} - X_t^{\text{joint}_2}
\end{equation}
\begin{equation}\label{eq: jm}
X_t^{\text{jm}} = X_t^{\text{joint}} - X_{t-1}^{\text{joint}}
\end{equation}
\begin{equation}\label{eq: bm}
   X_t^{\text{bm}} = X_t^{\text{bone}} - X_{t-1}^{\text{bone}}
\end{equation}

\subsection{Skeleton-specific Augmentations}

To improve the generalization ability and robustness of the model, skeleton-specific data enhancements are applied to the input data $\mathcal{X}$, denoted as $\mathcal{X}' = \mathcal{A}(\mathcal{X})$, where $\mathcal{A}$ represents the transformation function. These augmentation methods include \textbf{Normal Augmentations}, \textbf{Extreme Augmentations}, \textbf{Mixing Augmentations}, and \textbf{Viewpoint-Invariant Transformations}. Normal augmentations enhance data diversity through simple perturbations, while extreme augmentations further perform spatial and temporal transformations such as cropping, rotation, time reversal, and Gaussian noise addition. Mixing augmentations combine different samples to generate new data, and viewpoint-invariant transformations ensure the stability of skeleton data under varying observation angles.

\subsection{Skeletal Data Representation}
The representation of skeleton data is critical for subsequent feature extraction and modeling, denoted as $\mathcal{R} = \mathcal{R}(\mathcal{X}')$, where $\mathcal{R}$ is the representation function. The main data representation methods include Sequential Representation, Pseudo-Image Representation, Graph-Structure Representation and Token-based Representation.

\textbf{Sequential Representation} treats skeleton data as a time-series input and is particularly suited to RNN-based methods. The sequential representation can be formally defined as Eq.\ref{eq:Rseq}.

\begin{equation} \label{eq:Rseq}
    \mathcal{R}_{\text{seq}} = \{X_t \mid t = 1, 2, \dots, T\}
\end{equation}

\textbf{Pseudo-Image Representation} converts skeleton data into a two-dimensional image format by mapping the temporal dimension \(T\) and joint dimension \(V\) to the image height \(H\) and width \(W\), making it compatible with CNN-based models. The pseudo-image representation is given by Eq.\ref{eq:Rimg}.

\begin{equation} \label{eq:Rimg}
    \mathcal{R}_{\text{img}} = f_{\text{img}}(\mathcal{X}) \in \mathbb{R}^{T \times V \times C}
\end{equation}

\textbf{Graph-Structure Representation} models the skeleton as a graph \(G = (V, E)\), where \(V\) denotes joint nodes and \(E\) denotes bones connecting them, as shown in Eq.\ref{eq:Rg}.

\begin{equation} \label{eq:Rg}
    \mathcal{R}_{\text{graph}} = G = (V, E)
\end{equation}


\textbf{Token-based Representation} transforms each joint’s data into independent tokens via the embedding function \( TokenEmbed \), then applies position encoding \( PosEmbed \) to preserve and distinguish temporal and positional relationships between tokens. The token representation can be formally defined as Eq.\ref{eq:Rtoken}
, where \( \mathbf{z}_{t,v} \) is computed as Eq.\ref{eq:ztv}.
This representation is well suited to Transformer-based methods.

\begin{equation} \label{eq:Rtoken}
    \mathcal{R}_{\mathrm{token}}=\{\mathbf{z}_{t,v}\mid t=1,2,\ldots,T,v=1,2,\ldots,V\}
\end{equation}
\begin{equation} \label{eq:ztv}
    \mathbf{z}_{t,v}=TokenEmbed(X_t[v])+PosEmbed(t,v)
\end{equation}


These representations enable RNNs, CNNs, GCNs and Transformers to effectively model the structure and dynamics of skeleton data in subsequent stages.

\subsection{Feature Extraction}
Feature extraction is a crucial step in skeleton-based action recognition, aimed at extracting discriminative spatio-temporal features from the data to enhance recognition performance. Specifically, it includes Temporal Feature Extraction and Spatial Feature Extraction, denoted as $ \Phi(\mathcal{R}) \rightarrow  F_{\text{temporal}}, F_{\text{spatial}}$, where $\Phi$ represents the feature extraction function applied to the input data $\mathcal{R}$. \textbf{Temporal feature extraction} focuses on capturing the dynamic changes of the skeleton data over the temporal dimension, such as the speed and continuity of actions.\textbf{ Spatial feature extraction} emphasizes the static relationships, geometric arrangements, and structures among different joints or bones within the skeleton. Different network approaches, such as RNNs, CNNs, GCNs, and Transformers, exhibit distinct advantages at this stage, ensuring the effective extraction of key features relevant to the action categories.

\subsection{Spatial-Temporal Modeling}
Spatio-temporal modeling is the core step in skeleton-based action recognition, aiming to jointly model the spatial structure and temporal dynamics of skeleton data. It is expressed as $F_{\text{final}} = \Phi_{\text{st}}(F_{\text{temporal}}, F_{\text{spatial}})$, where $\Phi_{\text{st}}$ represents the spatio-temporal modeling method. Generally depending on the approach, spatio-temporal modeling falls into the following three main structures.

\textbf{Spatio-temporal serial structure}, as shown in Eq.\ref{eq:Fserial}, extracts spatial features first and then models the temporal relationships sequentially, making it suitable for staged processing.

\begin{equation} \label{eq:Fserial}
    F_{\text{serial}} = \Phi_{\text{temporal}}(\Phi_{\text{spatial}}(\mathcal{R}))
\end{equation}

\textbf{Spatio-temporal parallel structure}, as shown in Eq.\ref{eq:Fparallel}, simultaneously models spatial and temporal information to enable joint learning. 
\begin{equation} \label{eq:Fparallel}
    F_{\text{parallel}} = \Phi_{\text{temporal}}(\mathcal{R}) + \Phi_{\text{spatial}}(\mathcal{R})
\end{equation}

\textbf{Spatio-temporal fusion structure}, as shown in Eq.\ref{eq:Ffusion}, combines spatial and temporal features through flexible mechanisms, such as weighted summation, concatenation, or attention mechanisms, to enhance overall modeling performance. The fusion function $\text{Fusion}(\cdot) $ can adopt methods such as weighted summation, concatenation, or attention mechanisms.
\begin{equation} \label{eq:Ffusion}
    F_{\text{fusion}} = \text{Fusion}(\Phi_{\text{temporal}}(\mathcal{R}), \Phi_{\text{spatial}}(\mathcal{R}))
\end{equation}

Finally, the output prediction is performed based on a fully connected layer with parameters $W$ and $b$, and the predicted action category $\hat{y}$ is obtained using the softmax function as Eq.\ref{eq:softmax}.

\begin{equation}    \label{eq:softmax}
    \hat{y} = \text{Softmax}(W \cdot F_{\text{final}} + b).
\end{equation}

\section{Skeleton-based Modalities Generation}

\label{Modalitites}
Skeleton-based action recognition relies on skeleton sequences as the primary input data, which consist of a series of skeleton coordinates. Based on this raw joint coordinate data, it is natural to derive additional representations by defining various relationships within the skeleton structure. For instance, the edges formed between pairs of joints can serve as bone modality information, while planes defined by three joints can represent another type of modality. These multiple modalities are essentially supplementary data derived from skeleton sequences through predefined methods. Notably, many studies have demonstrated that different patterns of data exhibit distinct characteristics, which can complement each other in action recognition tasks. This section first reviews several classic approaches that leverage such modalities and then discusses the current mainstream paradigms of modality design based on deep learning techniques.

\subsection{Geometric Features}
Early research focused on recognizing actions by measuring skeleton parameters like angles, positions, orientations, velocities, and accelerations \cite{xia2012view, ohn2013joint, chaudhry2013bio, ofli2014sequence, muller2005efficient}. Chen et al. \cite{chen2010learning} identified nine geometric features and combined them to represent pose and motion. Yao et al. \cite{yao2012coupled} developed pose-based features, such as joint distances, joint-plane distances, and joint velocities. Ofli et al. \cite{ofli2014sequence} used angles between bones to select informative joints for classification. Vemulapalli \cite{vemulapalli2014human} used rotational relationships between body parts in Lie groups. These geometric features are intuitively understandable, describing relationships between joints, lines and planes.

\begin{figure}[t]
	\centering
	\includegraphics[width=1.0\linewidth]{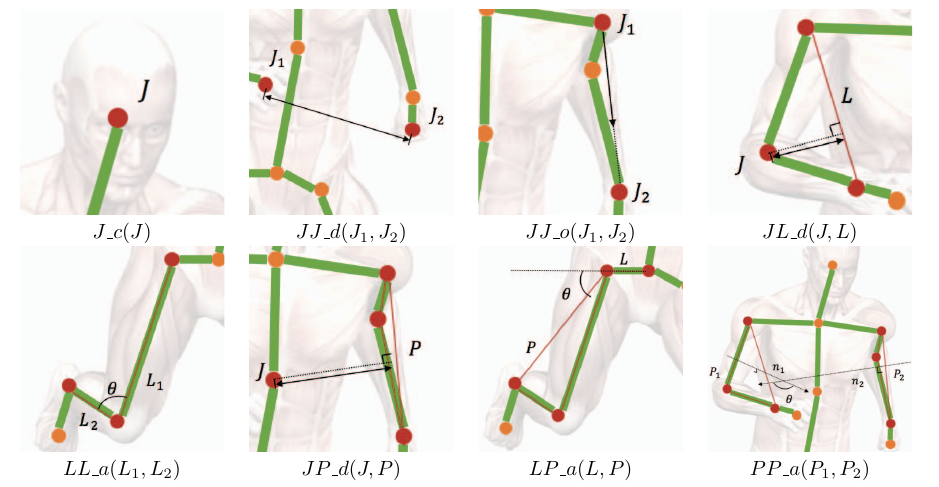}
	\caption{Designed eight geometric relational features: Joint Coordinate, Joint-Joint Distance, Joint-Joint Orientation, Joint-Line Distance, Line-Line Angle, Joint-Plane Distance, Line-Plane Angle, Plane-Plane Angle.}
	\label{fig:geofea}
	\vspace{-0.4cm}
\end{figure}

With the rise of deep learning, many works have extended geometric features for action recognition by employing various advanced deep learning methods. Huang et al. \cite{huang2017deep} extended \cite{vemulapalli2014human} by incorporating Lie group structure into a deep network, enabling it to effectively learn rotation-based features. Zhang et al. \cite{zhang2017geometric, zhang2018fusing} explored complex geometric relationships between joints, feeding rich spatial domain features into a three-layer LSTM network and designing eight relational features, as shown in Fig. \ref{fig:geofea}. Extensive experiments demonstrated that the joint-to-line distance feature outperformed other geometric features. Li et al. \cite{li2019learning} proposed view-invariant shape motion representations derived from skeleton sequences using geometric algebra, which significantly improved recognition accuracy. Liu et al. \cite{liu2020rotation} introduced a novel feature descriptor based on rotational relationships between joints in skeletons, leveraging geometric algebra to compute and derive these rotation relations using a specific operator known as the rotor in GA, thus offering more robust and meaningful features for action recognition.

\subsection{Generalized Features}
Although geometric features can outperform methods relying solely on joint coordinates, the diverse and intricate designs of these features often hinder the model's generalization performance. Some geometric features show clear advantages only on specific datasets. As a result, researchers have shifted their focus towards more generalized modal representations. The four predominant modalities namely Joint, Bone, Joint-Motion and Bone-Motion have gradually become mainstream.

\textbf{Bone.}
With the exception of original joints, many works \cite{zhang2017geometric,shi2019skeleton,shi2019two} have proved that the lines (bones) between pair-wise joints are also important geometric structures in the skeleton and contain rich structural or relational information. 

In particular, since each bone is connected to two joints, Shi et al. \cite{shi2019two} firstly defined the joint closest to the center of mass of the skeleton as the source joint, and the joint farther from the center of mass as the target joint. Each bone is then represented as a vector pointing from the source joint to the target joint, which captures both length and directional information. 
For instance, given a bone with its source joint $J_1 = (x_1, y_1, z_1)$ and its target joint $J_2 = (x_1, y_1, z_1)$, the bone vector is computed as Eq.\ref{eq:Bj1j2}.
\begin{equation} \label{eq:Bj1j2}
    B_{J1,J2} = (x_2 - x_1, y_2 - y_1, z_2 - z_1)
\end{equation}

Specifically, joints emphasize the absolute position, which can figure out discriminative moving parts of body in the action by analyzing the distribution or local density of joints. While bones emphasize relative position, which can build angles to help figure out specific poses and actions. Thus, there exists potential complementarity for action recognition between both geometric structures.

\textbf{Temporal Motion.}
Inspired by the work of Simonyan and Zisserman \cite{simonyan2014two}, where RGB images and optical flow fields are utilized as two input streams of a network, Li et al. \cite{li2017skele} extended this concept by defining temporal motion extracted from skeleton data as an additional input. Specifically, given a 3D joint coordinate $\boldsymbol{J} = (x, y, z)$, the skeleton of an individual is represented as a set of joint coordinates $\boldsymbol{X}_t=\{\boldsymbol{J}_t^1,\boldsymbol{J}_t^2,\ldots,\boldsymbol{J}_t^N\}$ where $N$ denotes the number of joints in the skeleton for each frame. The skeleton motion between two consecutive frames is computed as Eq.\ref{eq:mt}.
\begin{equation} \label{eq:mt}
\begin{aligned}
    \boldsymbol{M}_t
    & =\boldsymbol{X}_{t}-\boldsymbol{X}_{t-1} \\
    & =\{\boldsymbol{J}_{t}^1-\boldsymbol{J}_{t-1}^1,\boldsymbol{J}_{t}^2-\boldsymbol{J}_{t-1}^2,\ldots,\boldsymbol{J}_{t}^N-\boldsymbol{J}_{t-1}^N\}
\end{aligned}
\end{equation}

where $t$ is the frame index. Temporal motion, also referred to as joint-motion, is informative in discriminating fine-grained actions, such as "put on jacket" versus "take off jacket." Temporal motion can also be computed for the bone stream in the same manner as joint motion, yielding bone-motion.

\textbf{Mainstream Trends.} 
Different types of data exhibit distinct characteristics, and fusing multiple modalities can enhance performance. Leveraging the simple yet effective modal designs of Joint, Bone, Joint-Motion and Bone-Motion, an increasing number of studies \cite{zheng2019relational,liu2020msg3d,li2021symbiotic,chen2021channel,chi2022infogcn,qiu2022spatio,liu2023painet,do2025skateformer} have adopted these four modalities as inputs for modeling. Specifically, most models are trained using these four modalities separately, with results obtained by averaging the outputs from each modality.

\section{Skeleton-specific Augmentations}
\label{augmentations}

Most deep learning methods require large amounts of data to achieve optimal performance. However, most existing skeleton-based action recognition algorithms have focused on network architectures \cite{ren2024survey,sun2022human}, and one critical issue that has received little attention is the inadequacy of skeleton data. The lack of annotated training data often leads to overfitting, reducing the model's generalization ability. Therefore, leveraging data augmentation techniques to expand limited skeleton data can enhance both model performance and generalization. This process is essential in the data preprocessing stage before model development. Data augmentation requires distinct techniques depending on the modality and task. For instance, image and video data may use techniques like dropout \cite{srivastava2014dropout,tompson2015efficient}, image rotation \cite{yang2022image}, random cropping \cite{yang2022image}, occlusion \cite{zhong2020random}, and jittering \cite{krizhevsky2012imagenet}, while natural language processing may use noise-based methods \cite{wei2019eda,li2022dataaugnlp}.

On the one hand, skeleton sequences are non-Euclidean data, so augmentation methods used for images and videos cannot be directly transferred to them. On the other hand, skeleton sequences exhibit minimal redundancy and carry a high degree of semantic information. Consequently, to effectively learn representations for 3D skeleton data, standard RGB-based augmentations, such as color distortion or Gaussian blurring, are inadequate. Instead, skeleton-specific augmentation strategies are required to ensure the learned features capture the spatiotemporal dynamics of the joints.

\begin{figure}[t]
	\centering
	\includegraphics[width=1.0\linewidth]{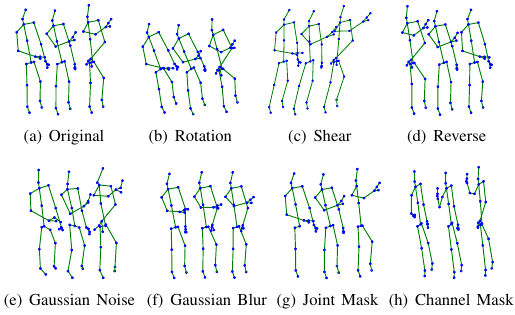}
	\caption{Visualization of normal augmentations.}
	\label{fig_aug_normal}
	\vspace{-0.4cm}
\end{figure}

\subsection{Normal Augmentations}
Normal Augmentations \cite{rao2021augmented,thoker2021skecontra}, an early approach to data augmentation, include techniques such as Spatial Flip, Rotation, Gaussian Noise, Gaussian Blur, and Channel Masking. This paradigm leverages the pattern invariance in skeleton sequences, encouraging models to learn intrinsic action patterns across various transformations. Due to its robustness, this approach has been widely adopted in skeleton-based action recognition models \cite{yan2018stgcn,cheng2020shiftgcn,liu2020msg3d,shi2019two,chen2021channel,chi2022infogcn,xin2023mixformer,wang2017modeling}. Yan et al. \cite{yan2018stgcn} used random moving and sampling during training to augment skeleton data and mitigate overfitting. Subsequent works \cite{cheng2020shiftgcn,liu2020msg3d,shi2019two} used similar techniques. Since Zhang et al. \cite{zhang2020semantics} proposed random rotation of 3D skeletons at the sequence level, state-of-the-art methods \cite{chen2021channel,chi2022infogcn,xin2023mixformer} have adopted this technique.

In addition, in self-supervised tasks based on contrastive learning, numerous studies \cite{rao2021augmented,thoker2021skecontra,lin2025idempotent,hua2023part,zhou2023pstl,franco2023hyperbolic,shah2023halp,wu2024scd} have explored innovative skeleton data augmentation strategies to enhance contrastive learning. These augmentations drive the model to learn intrinsic action patterns by leveraging diverse skeleton transformations, further improving its ability to capture meaningful representations. Rao et al. \cite{rao2021augmented} proposed innovative skeleton data augmentation strategies to generate query and key skeleton sequences as augmented instances for contrastive learning. They introduce seven augmentation techniques, including rotation, shearing, flipping, Gaussian noise, Gaussian blur, joint masking, and channel masking. These methods aim to simulate diverse viewpoint changes and noise conditions through random transformations and data perturbations, thereby enhancing the model's robustness for action recognition and enabling it to capture intrinsic action patterns. Similarly, Thoker et al. \cite{thoker2021skecontra} presented a suite of spatial and temporal skeleton augmentation techniques to generate positive pairs for 3D skeleton-based action sequences. Their approach includes pose augmentation, joint jittering, and temporal crop-resize, which are combined to form a comprehensive spatio-temporal skeleton augmentation pipeline. This strategy encourages the model to focus on the spatiotemporal dynamics of skeleton-based action sequences while mitigating the influence of confounding factors such as viewpoint variations and exact joint positions.

\subsection{Extreme Augmentations}
\begin{figure}[t]
	\centering
	\includegraphics[width=1.0\linewidth]{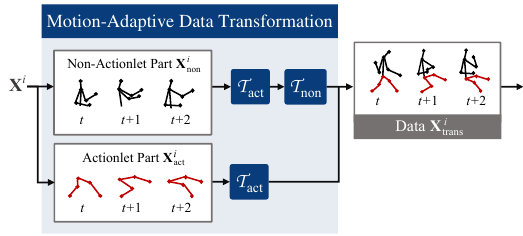}
	\caption{The pipeline of motion-adaptive data transformation.}
	\label{fig_actionlet}
	\vspace{-0.4cm}
\end{figure}

However, those normal augmentations limit the encoder from exploring novel patterns exposed by other augmentations. Extreme augmentations \cite{guo2022contra} introduce more movement patterns for learning general feature representations, including four spatial augmentations: Shear, Spatial Flip, Rotate, Axis Mask, two temporal augmentations: Crop, Temporal Flip, and two spatio-temporal augmentations: Gaussian Noise and Gaussian Blur. While normal augmentations show robustness, they lack effective design to fully leverage strong augmentations, leaving their utility underexplored. To address this, Zhang et al. \cite{zhang2023HiCLR} proposed a progressive strategy generating ordered positive pairs through incremental augmentation, incorporating strong augmentations to enhance representation learning. 
Balancing the extraction of rich augmented patterns with the preservation of motion information is crucial, particularly when dealing with extreme augmentations. 
To address this, Lin et al. \cite{lin2023actionlet} proposed motion-adaptive data transformations based on actionlets, applying extreme augmentations to non-actionlet regions, as shown in Fig. \ref{fig_actionlet}. These methods enhance the model's ability to learn intrinsic patterns and improve the effectiveness of contrastive learning.

\subsection{Mixing Augmentations}
Current methods for normal \cite{rao2021augmented,yan2018stgcn,zhang2020semantics} or extreme \cite{guo2022contra,zhang2023HiCLR} augmentations primarily apply global rotations or slight perturbations to skeleton data, ensuring that the semantic information remains intact during the data augmentation process. Given the inherent challenge in augmenting individual skeleton data samples, a feasible approach is to consider data augmentation across multiple samples. In recent years, various mix-based data augmentation methods \cite{hendrycks2022pixmix,liu2022automix,qin2020resizemix,verma2019manifold,yun2019cutmix,zhang2017mixup} have emerged in domains such as image classification, fine-grained image recognition, graph classification, and object detection, aiming to improve the generalization capabilities of neural networks and performance in other complex tasks. Therefore, inspired by the success of mix-based data augmentation techniques in other domains, several studies \cite{zhang2023pcm3,zhang2024shapmix,xiang2024jointmix} have proposed mixing-based augmentation methods tailored for skeleton-based action recognition. 

Zhang et al. \cite{zhang2023pcm3} pioneered the application of mix-based augmentation methods to skeleton data, introducing techniques such as CutMix \cite{yun2019cutmix}, ResizeMix \cite{qin2020resizemix}, and Mixup \cite{zhang2017mixup} as inter-skeleton transformations to generate mixed skeleton datasets. Building on this, they addressed the challenge of long-tailed distributions prevalent in deep learning by proposing a spatial-temporal mixing strategy \cite{zhang2024shapmix}. This approach explicitly links critical motion patterns with high-level semantics, facilitating more robust decision boundary learning for minority classes. Similarly, Xiang et al. \cite{xiang2024jointmix} proposed Joint Mixing Data Augmentation (JMDA) for skeleton-based action recognition, incorporating TemporalMix (TM) and SpatialMix (SM). Recently, to facilitate stable training of transformer-based models, which typically require large amounts of data, Do et al. \cite{do2025skateformer} proposed a comprehensive set of augmentation techniques. These include intra-instance augmentations, which apply standard transformations, and inter-instance augmentations, which involve mixing data across instances. These methods aim to enhance model performance by mitigating overfitting and improving generalization, effectively addressing the limitations posed by the scarcity of skeleton data.


\subsection{Viewpoint-Invariant Transform}

Human actions are often observed from arbitrary camera viewpoints in real-world scenarios, which results in some datasets containing data captured from multiple perspectives. One of the primary challenges in skeleton-based human action recognition lies in the complex viewpoint variations inherent in the data. Skeletal representations of the same posture can vary significantly when viewed from different angles, making action recognition a particularly difficult task in practice \cite{aggarwal2014human,ji2009advances}. To better mitigate the challenges posed by viewpoint variations, it is often necessary to preprocess the raw skeleton data before converting it into a structured format.

One common approach to addressing this challenge of viewpoint variations is to apply a preprocessing step that transforms 3D joint coordinates from the camera's coordinate system to a person-centered coordinate system.
This is typically achieved by aligning the body center to the origin and rotating the skeleton so that the body plane is parallel to the \( (x, y) \) plane, thereby making the skeletal data invariant to absolute position and body orientation \cite{zhu2016colstm,shahroudy2016ntu,liu2016spatio,song2017end}. 
For instance, Lee et al. \cite{lee2017ensemble} converted the original skeletal sequence into a human-perception-based coordinate system, mitigating confusion caused by inconsistent orientations in the skeletal sequence and improving robustness to scaling, rotation, and translation of skeletal data.

However, the approaches mentioned above have notable limitations.
First, the method relies on defining the body plane using joints such as the "hip", "shoulder" or "neck" which may not always suit alignment purposes due to the non-rigid nature of the human body. Second, these transformations can result in the loss of critical motion information, such as the trajectory and speed of the body center or the dynamic changes in body orientation. For example, the action of walking may be reduced to walking in place, and a dance involving body rotation may appear as dancing with a fixed orientation. Finally, the preprocessing steps (e.g., translation and rotation) are designed based on human-defined heuristics rather than being optimized explicitly for action recognition tasks, limiting the potential to exploit optimal viewpoints effectively. Hence, Zhang et al. \cite{zhang2017view} designed a view adaptive recurrent neural network (RNN) with LSTM architecture, which enables the network itself to adapt to the most suitable observation viewpoints from end to end. Friji et al. \cite{friji2021geometric} proposed a geometry-based deep learning method, which optimizes rigid and non-rigid transformations to filter out the effects of position, scale, and rotation.


\section{Skeletal Data Representation}
Generally, data augmentation techniques are directly applied to the unstructured raw skeleton sequences, which are typically in a form that models cannot directly interpret. To enable models to effectively learn from skeleton data, the next step involves converting the raw skeleton data into a format compatible with input requirements for RNNs, CNNs, GCNs or Transformers. In this process, the key challenge is to efficiently transform the sparse, unstructured skeleton action sequences into a structured format that deep learning models can understand while preserving as much of the original spatial and temporal dependencies as possible.
In particular, \textbf{graph-structured representation} with GCNs convert the skeleton into a graph that mirrors the human anatomical topology and will be introduced in greater detail.


\subsection{Raw Skeleton Sequences: 2D vs. 3D}
Skeleton data can be represented in both 2D and 3D formats, and both forms are capable of supporting action recognition tasks \cite{avola20192,elias2019understanding}. 
However, compared to 2D skeletons, 3D skeletons provide additional depth information, which significantly enhances the preservation of spatial details that are crucial for the quality of subsequent modeling. Furthermore, with the advancement of depth image sensors, inertial sensors, and human pose estimation algorithms, obtaining 3D skeleton data has become increasingly easier. As a result, 3D skeleton-based benchmarks have gained greater popularity in human action recognition tasks, and methods relying on 3D skeleton data are now more prevalent. 
Unless otherwise stated, this paper primarily focuses on 3D skeleton data.

\begin{figure}[t]
	\centering
	\includegraphics[width=1.0\linewidth]{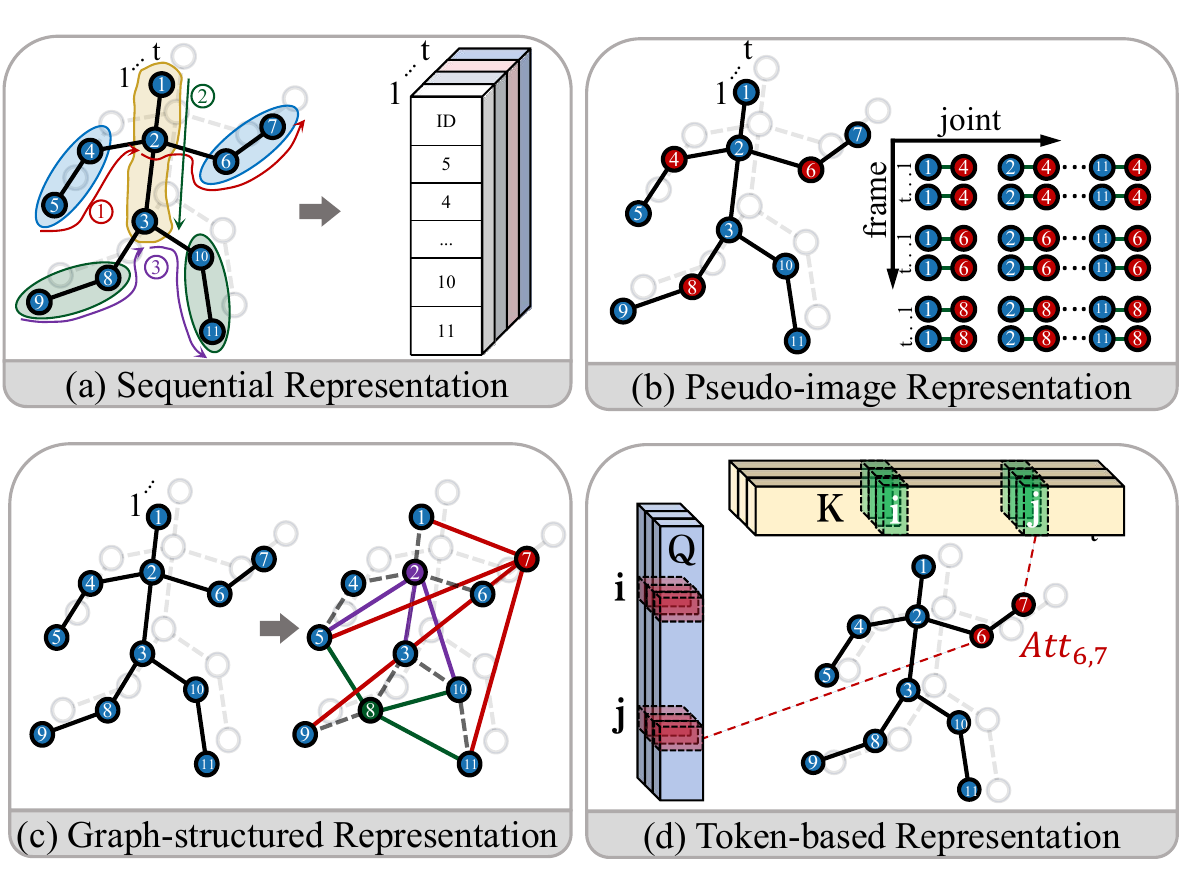}
	\caption{Unstructured skeleton sequences are converted into four distinct input formats for further processing.}
	\label{fig:repre}
	\vspace{-0.4cm}
\end{figure}

\subsection{Sequential Representation with RNNs}

In the early stages of research, models such as standard RNNs, LSTMs, and GRUs have demonstrated significant effectiveness in processing sequential data \cite{liu2016tree,liu2017tree,wang2017modeling,avola20192p,zheng2019relational,zhang2017geometric,zhang2018fusing,lee2017ensemble,zhang2017view,friji2021geometric}. Fig.\ref{fig:repre} (a) illustrates interpreting the skeleton sequence as a sequential representation.
Consequently, a critical initial consideration is how to effectively transform skeletal data into a sequential format suitable for these models while preserving the spatiotemporal information inherent in the original skeletal data. 
However, most early works simply arrange joints in a linear sequence, overlooking the kinematic dependencies between them. This approach introduces false connections between body joints that are not strongly related, leading to inaccuracies in modeling the human body’s structure.
Liu et al. \cite{liu2016tree,liu2017tree} considered the natural structure and spatial relationships of human joints by modeling them as tree-structured sequential representations. This approach effectively preserves the inherent hierarchical and adjacency relationships between joints, thereby capturing richer spatial dependency features. 
Similarly, Wang et al. \cite{wang2017modeling} proposed two methods for converting the skeleton structure into a sequence before applying an RNN to capture spatial dependencies. One of these methods is graph traversal-based, where joints are accessed sequentially based on their adjacency relationships, ensuring that the spatial structure of the graph is preserved.

\subsection{Pseudo-image Representation with CNNs}
The representation of skeleton data for CNN-based methods has evolved significantly to effectively encode spatiotemporal information. Fig.\ref{fig:repre} (b) illustrates interpreting the skeleton sequence as a pseudo-image representation.
To meet the input requirements of CNNs, 3D skeleton sequence data is often transformed from a vector sequence into a pseudo-image. However, creating a representation that captures both spatial and temporal information can be challenging. To address this, many researchers have encoded skeleton joints into multiple 2D pseudo-images, which are then fed into CNNs for efficient feature learning \cite{ding2017investigation, xu2018ensemble,wang2018action,li2017skeleton,li2019learning,liu2017enhanced,caetano2019skelemotion,li2019learning}.

\begin{figure}[t]
	\hfill
	\begin{minipage}[b]{1.0\linewidth}
		\centering
		\centerline{\includegraphics[width=1.0\linewidth]{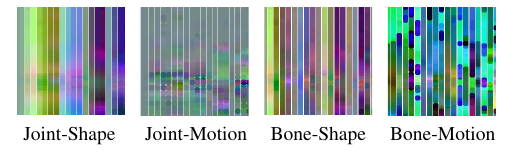}}
		\vspace{-0.1cm}
		\centerline{(a)}\medskip
	\end{minipage}
	\hfill
	\begin{minipage}[b]{1.0\linewidth}
		\centering
		\centerline{\includegraphics[width=1.0\linewidth]{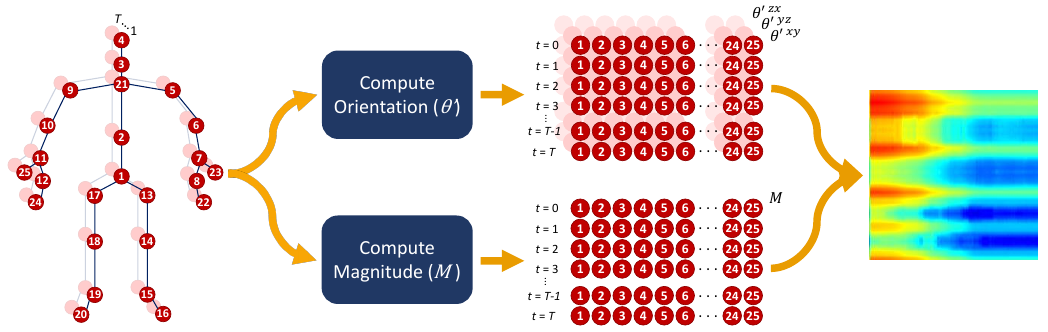}}
		\vspace{-0.1cm}
		\centerline{(b)}\medskip
	\end{minipage}
	\vspace{-0.5cm}
	\caption{Examples of the proposed representation skeleton image. (a) shows Skeleton sequence shape-motion representations generated from "pick up with one hand" action. (b) shows the SkeleMotion representation workflow.}
	\label{fig_repre_cnn}
	\vspace{-0.4cm}
\end{figure}

Wang \cite{wang2018action} proposed the Joint Trajectory Maps (JTM), which represent spatial configuration and dynamics of joint trajectories into three texture images through color encoding. However, this method is a little complicated and also loses importance during the mapping procedure. To tackle this shortcoming, Li et al. \cite{li2017skeleton} used a translation-scale invariant image mapping strategy, which first divided human skeleton joints in each frame into five main parts according to the human physical structure, and then those parts were mapped to 2D form. This method makes the skeleton image consist of both temporal and spatial information. However, though the performance was improved, there is no reason to treat skeleton joints as isolated points, because in the real world, an intimate connection exists among our body parts. For example, when waving the hands, not only the joints directly within the hand should be considered, but also other parts such as shoulders and legs are important. Li et al. \cite{li2019learning} proposed the shape-motion representation from geometric algebra, which addressed the importance of both joints and bones and fully utilized the information provided by the skeleton sequence. Similarly, Liu et al. \cite{liu2017enhanced} also used the enhanced skeleton visualization to represent the skeleton data, and Caetano et al. \cite{caetano2019skelemotion} proposed a new representation named SkeleMotion based on motion information, which encodes the temporal dynamics by explicitly computing the magnitude and orientation values of the skeleton joints. Fig. \ref{fig_repre_cnn}(a) shows the shape-motion representation \cite{li2019learning}, while Fig. \ref{fig_repre_cnn}(b) illustrates the SkeleMotion representation. 
Moreover, similar to SkeleMotion, Caetano et al. \cite{caetano2019skeleton} adopted the SkeleMotion framework, but based it on a tree structure and reference joints for skeleton image representation.
Previous methods usually utilize body joint-based representation, while leaving edge-based movement poorly investigated. The movement of the skeleton edge provides richer information than body joints for recognizing a particular action. Some human actions, such as playing the guitar, involve repeatedly swinging one part of the body, which is not easy to identify from the perspective of motion because the coordinates of the body joints do not change much. Wang et al. \cite{wang2021skeleton} proposed the novel skeleton edge motion networks (SEMN) to further explore the motion information of human body parts. Specifically, SEMN addresses the movement of the skeleton edge by using the angle changes of the skeleton edge and the movement of the corresponding body joints.

\subsection{Graph-structured Representation with GCNs}
Inspired by the fact that human 3D skeleton data naturally forms a topological graph, rather than a sequence vector or pseudo-image as used in RNN-based or CNN-based methods, Graph Convolutional Networks (GCNs) have recently been adopted for skeleton-based action recognition. Fig.\ref{fig:repre} (c) illustrates interpreting the skeleton sequence as a graph-structured representation. This approach leverages the inherent graph structure of the data \cite{yan2018stgcn, li2019actional, shi2019two, zhang2020semantics, ye2020dynamic}. The core idea is to model the skeleton as a spatiotemporal graph, where nodes represent joints and edges encode relationships both within the body and across time. This representation has demonstrated compelling results, further validated by strong performance on benchmark datasets. Specifically, in practical implementation, the skeleton topology is represented as an adjacency matrix. For clarity, we use this adjacency matrix to depict the connections between human body joints in this section. By repeatedly multiplying the adjacency matrix $l$ times, we obtain a high-order polynomial of the adjacency matrix. The $l$-th order polynomial reflects the aggregation of information from $l$-hop neighbor nodes. Structural links are derived by combining polynomials of different orders in varying proportions.

\textbf{Fixed Topological Representation.}
The most straightforward approach is to construct a fixed adjacency matrix based on the topological connectivity between joints to represent the graph structure. However, using a naturally connected skeleton topology as the adjacency matrix may not always be the optimal choice. Consequently, considerable effort has been dedicated to optimizing the structure of the skeleton topology.

Yan et al. \cite{yan2018stgcn} first introduced a novel model for skeleton-based action recognition, called Spatial-Temporal Graph Convolutional Networks (ST-GCN). In this approach, they constructed a spatiotemporal graph, where the joints are represented as graph vertices, and the natural connectivity within the human body structure and across time serves as graph edges. 
The adjacency matrix partitioning strategy used in ST-GCN divides the matrix into three parts: the center node itself, nodes close to the center of gravity, and nodes further away from the center. This partitioning scheme restricts nodes from interacting only with their immediate neighbors, which fails to capture the broader interactions of many nodes in human movement. 
To address this limitation, Wang et al. \cite{wang2022skeleton} extended the adjacency matrix into five parts, incorporating two additional sets of nodes—one near the central node and the other farther away. This modification enables each node to interact with those up to two hops away, thus providing a more comprehensive representation of long-term and global motion characteristics.
To capture richer dependencies, Li et al. \cite{li2019actional} introduced Action-Structural Graph Convolutional Networks (AS-GCN). In addition to using Actional Links to model specific latent dependencies related to actions, they also extended the existing skeleton graph to represent higher-order dependencies, known as Structural Links. These two types of links are combined into a generalized skeleton graph, which improves feature learning and enables the model to construct a more generalized representation of the skeleton. 

\begin{figure}[t]
	\centering
	\includegraphics[width=1.0\linewidth]{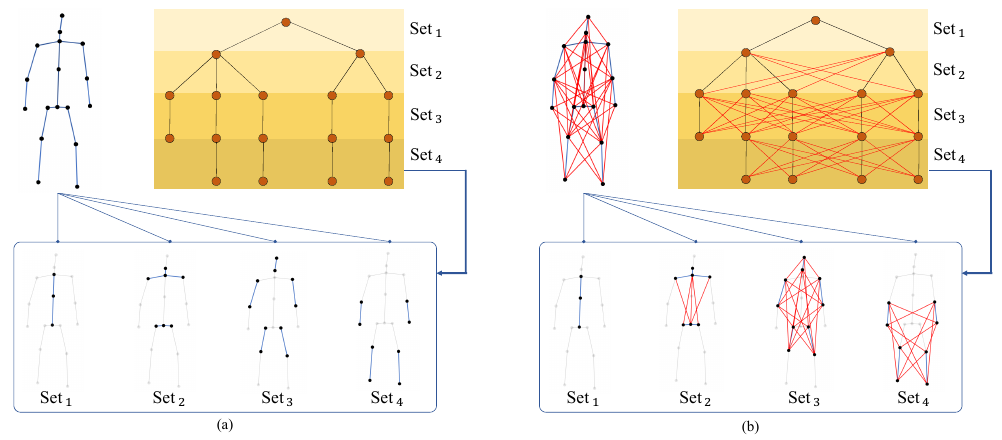}
	\caption{Structure of the HD-Graph with physically connected edges and fully connected edges.}
	\label{fig:repgcn2}
	\vspace{-0.4cm}
\end{figure}

However, approaches mentioned above may introduce noise from uncorrelated nodes, which can negatively affect the neural network. 
To mitigate this issue and capture dependencies between non-physically connected joints, Hao et al. \cite{hao2021hypergraph} proposed Hyper-GNN. In this method, hypernodes are selected from the top 50$\%$ of joints with the longest movement distances within the video sequence. Hyperedges connect these distant, non-adjacent nodes, enabling Hyper-GNN to extract complex higher-order spatial-temporal features.
Similarly, Lee et al. \cite{lee2022hierarchically} introduced HD-GCN, which decomposes the human skeleton graph into six hierarchical subsets to emphasize the distance relationships between joints. The HD-Graph structure, shown in Fig. \ref{fig:repgcn2}, includes both physically connected edges and fully connected edges, allowing nodes that are not naturally connected to be linked within subgraphs. This improves the transmission of information between joints with hierarchical distances. 
Qin et al. \cite{qin2022multi} proposed a method that down-samples the original skeleton topology into six compressed topologies in stages. Each compression step aggregates higher-order neighbor information, enhancing GCN’s ability to identify relationships between joints at varying distances.


\textbf{Dynamic Topological Representation.}
However, the skeleton topology graph used by ST-GCN is handcrafted and fixed, limiting the neural network to learning only the relationships between naturally connected joints. This constraint impedes the ability of GCNs to fully extract and aggregate relevant features from the skeleton data. Another fundamental challenge with GCNs is their limited capacity to model long-term dependencies between distant joints. For example, in the 'clapping' action, the joints on the two palms are spatially distant in the skeleton topology, making it difficult for the graph convolution operation to capture the relationships between these joints. These challenges in GCNs can be summarized as: \textbf{(1) how to extract richer feature representations from the skeleton topology, and (2) how to model long-term dependencies between joints.} Addressing these issues calls for more dynamic topological representations of skeleton data.

To extract and learn richer feature representations, Shi \cite{shi2019two} proposed 2s-AGCN, which constructs an adaptive topology graph. This graph is updated automatically via the backpropagation algorithm, enabling it to more effectively capture the joint connection strength. Specifically, the adaptive adjacency matrix consists of three components: (1) an adjacency matrix representing the physical connections between body joints, (2) a purely parameterized adjacency matrix that is jointly optimized with the network's other parameters during training, and (3) a sample-specific adjacency matrix tailored to each individual input sample.
Ye et al. \cite{ye2020dynamic} proposed Dynamic GCN, which utilizes a context-encoding network to model a dynamic adjacency matrix that incorporates global contextual information. The context-encoding network consists of three \(1 \times 1\) convolutional layers, which compress the skeleton feature maps along both the channel and temporal dimensions, and then reshape them to match the size of the adjacency matrix.
Chen et al. \cite{chen2021channel} proposed the CTR-GCN which models channel-wise topologies by dividing the channels into groups and only sharing the skeleton topology between groups. CTR-GCN learns a group-shared topology and channel-specific correlations simultaneously to enhance the feature extraction and aggregation ability of GCNs without significantly increasing the parameters. 
Liu et al. \cite{liu2023tsgcnext} further proposed Dynamic-Static Multi-Graph Convolution (TSGCNeXt) which expands the channel number of skeleton features by four times, mapping skeleton features into higher-dimensional feature spaces to capture discriminative potential spatial dependencies.

\subsection{Token-based Representation with Transformers}
In recent years, with the impressive performance of Transformers across various research fields, many Transformer-based methods have emerged \cite{cho2020self, shi2020decoupled, plizzari2021spatial, zhang2021stst, chen2022hierarchically}. Fig.\ref{fig:repre} (d) illustrates interpreting the skeleton sequence as a token-based representation. The Transformer architecture is built on the Self-Attention mechanism, which receives input as a 1D sequence of token embeddings and computes the relationships between all input vectors. However, since the Transformer model itself does not inherently handle sequential data, it is necessary to apply position encoding to the input skeleton data to provide spatial information about the joints within the sequence. Firstly, the input joints are weighted with additional information through position embedding to preserve spatial relationships. The most common approach in skeleton data processing employs absolute position encoding, where each joint is labeled using sine and cosine functions with varying frequencies as the encoding functions, as Eq.\ref{eq:PE}.

\begin{equation} \label{eq:PE}
    \boldsymbol{PE(p, i)} =
\begin{cases} 
\sin\left(\frac{p}{10000^{i/C_{in}}}\right) & \text{if } i \text{ is even,} \\
\cos\left(\frac{p}{10000^{i/C_{in}}}\right) & \text{if } i \text{ is odd.}
\end{cases}
\end{equation}

In Eq.\ref{eq:PE}, $p$ and $i$ denote the position of a joint and the dimension of its position encoding (PE) vector, respectively, the resulting PE vector is then added to the original joint vector. This ensures each joint is assigned a unique identifier. Subsequently, the PE vectors are projected into the embedding subspace through a mapping operation, akin to the word embeddings in NLP or the patch embeddings in VIT\cite{dosovitskiy2020image}.

\section{Feature Extraction}
In the field of skeleton-based human action recognition, feature extraction is one of the most crucial core links of the entire process. This process not only requires accurately capturing the temporal dynamic changes in the skeleton sequence but also pays full attention to the spatial relationship between joint nodes and their connection topology to facilitate subsequent spatio-temporal feature modeling. Recent methods have also begun to apply Large Language Models (LLMs), Generative Models, and other advanced learning methods to the feature extraction stage. These methods can assist the system in automatically mining deeper and more abstract action patterns, thereby significantly enhancing the model's learning ability and generalization performance. The main feature extraction method is shown in Fig. \ref{fig:Feature}.

\subsection{Temporal Feature Extraction}

\textbf{Temporal Feature Extraction Using RNNs.}\label{rnn-t}
RNN and its variants (e.g., LSTM) are widely used in skeleton-based action recognition due to their strong temporal sequence processing capabilities. Many studies focus on optimizing RNN and LSTM models for improved temporal feature extraction.  
Kim et al. \cite{veeriah2015differential} proposed dRNN, a model designed to capture salient spatio-temporal information by quantifying information gain changes caused by significant motion between frames. Zheng et al. \cite{zheng2019relational} developed an attention-based recurrent relational LSTM network that employs a multi-layer LSTM structure to capture temporal features in skeleton sequences.  
Liao et al. \cite{liao2021logsig} introduced the mathematical tool of log-signature into RNN, creating the Logsig RNN model. This approach significantly reduces the number of time steps required by RNNs while enabling the model to handle variable-length time series without padding, offering robustness against data loss.

\textbf{Temporal Feature Extraction Using CNNs.}
Temporal feature extraction poses a significant challenge for convolutional neural networks (CNNs) in skeleton-based action recognition.
Kim et al. \cite{soo2017interpretable} utilized Temporal Convolutional Networks (TCN)  \cite{lea2017temporal}, providing a clear and interpretable spatio-temporal representation. Liu et al. \cite{liu2019skepxels} proposed Skepxel, which hierarchically captured fine-grained temporal relationships among joints within frames. In contrast, Rahmani et al. \cite{rahmani20163d} proposed  Fourier Temporal Pyramid, emphasizes capturing coarse-grained temporal relationships for a broader temporal perspective.
To further enhance spatio-temporal feature representation, Jia et al. \cite{jia2020two} proposed TS-TCN, which integrated inter-frame and intra-frame vector features for improved sensitivity to skeletal dynamics compared to traditional distance- and angle-based features. Inter-frame vector features represent the motion of skeletal joints between consecutive frames, while intra-frame vector features capture the relative positions of joints within a single frame. Moreover, TS-TCN redesigns residual blocks with varying strides along the network depth, enhancing TCN’s ability to model long-term dependencies in complex action sequences effectively.

\textbf{Temporal Feature Extraction Using Temporal Graphs.}
Graph convolutional networks (GCNs) have been shown to excel in modeling skeleton space, and by incorporating temporal information into graph construction, dynamic changes in the temporal dimension can be modeled. This method usually builds a spatio-temporal graph based on skeleton sequence data, where joints are treated as graph vertices and the spatial and temporal relationships between joints are represented as graph edges.
Yan et al. \cite{yan2018spatial}  first introduced the concept of spatio-temporal graph and proposed the spatio-temporal graph convolutional network (ST-GCN). In this model, the joints in the skeleton sequence are defined as graph vertices, the spatial connections are derived from the human body structure, and the temporal connections are captured through inter-frame edges, enabling joint modeling of spatial and temporal features. This framework significantly enhances the network's ability to represent spatio-temporal features. Li et al. \cite{li2019spatio}further refined the construction of temporal graphs by introducing temporal graph routers. By feeding the trajectories of skeleton sequence nodes into this router, they generate a more flexible and adaptive skeleton joint connection graph. This method effectively improves the accuracy of temporal feature extraction and enhances the model's ability to capture complex motion patterns. Furthermore, Shift-GCN\cite{cheng2020shiftgcn} adopts a new method that combines shift map operations with point-wise convolutions. This technology provides a more flexible receptive field for temporal graphs, optimizes the aggregation of temporal features, and further enhances spatio-temporal graph modeling.
Graph Convolutional Networks (GCNs) are highly effective in modeling skeleton structures, representing skeleton sequence data as a spatio-temporal graph in which the joints serve as graph vertices, and the spatial-temporal relationships are represented by graph edges.
Yan et al. \cite{yan2018spatial} first introduced the spatio-temporal graph concept and proposed the spatio-temporal graph convolutional network (ST-GCN). 
Li et al. \cite{li2019spatio} refined temporal graph construction by introducing temporal graph routers, which adaptively generate skeleton joint connections using skeleton sequence trajectories. 
Cheng \cite{cheng2020shiftgcn} combined shift map operations with point-wise convolutions, providing a more flexible receptive field for temporal graphs. 

\textbf{Temporal Feature Extraction Using Attentions.}
Unlike traditional methods, attention mechanisms capture dependencies between time points globally, offering advantages in processing long sequences and complex spatio-temporal interactions.
Cho et al. \cite{cho2020self} proposed three self-attention networks (SAN) to extract long-term temporal semantic features by capturing long-range correlations through self-attention.  
Plizzari et al. \cite{plizzari2021spatial} designed a spatio-temporal Transformer network (ST-TR) that learned motion dynamics between frames through a temporal self-attention module, capturing temporal and spatial dependencies for comprehensive action representation. 
Zhang \cite{zhang2021stst} proposed a space-time dedicated Transformer (STST) that integrated a direction-time Transformer block with a direction-aware strategy to model long-term dependencies effectively, while leveraging multi-task self-supervised learning to enhance robustness against skeleton data noise and improve temporal feature extraction.
In addition to explicit temporal feature extraction, Wu et al. \cite{wu2024freq} introduced a frequency-aware attention module leveraging discrete cosine transform (DCT) for frequency-based temporal feature representation.
However, attention mechanisms require significant storage for global information, leading to high computational costs. Shi et al. \cite{shi2021star} addressed this with STAR, a sparse Transformer using piecewise linear attention for efficient temporal dynamics and variable frame-length support.

\textbf{Temporal Feature Extraction Hierarchically.}
Hierarchical temporal feature extraction, which decomposes dynamics into components with varied receptive fields, captures multi-scale patterns in skeletal sequences, offering a comprehensive temporal understanding of motion.
Lee et al. \cite{lee2017ensemble} proposed TS-LSTM, integrating short-, medium-, and long-term networks to capture multi-scale motion features.
Wang et al. \cite{wang2018skeleton} used a multi-stream LSTM to process single-frame, short-term, and long-term features for a holistic view. 
To address non-adjacent temporal dependencies, Liu \cite{liu2020disentangling} proposed a multi-scale temporal convolution module with diverse receptive fields, outperforming local convolutions. 
Duan \cite{duan2022dg} further improved efficiency by integrating channel grouping. 
Shi et al. \cite{shi2020decoupled} developed DSTA-Net, decoupling skeleton data into slow and fast temporal streams.
Chen et al. \cite{chen2022hierarchically} proposed Hi-TRS, a hierarchical Transformer framework that extracts spatio-temporal features at frame, clip, and video levels, with hierarchical supervision during pretraining.

\textbf{Temporal Feature Extraction by Partition.}
Directly inputting all information without segmentation often leads to coarse and imprecise temporal feature extraction. To address this, researchers have explored dividing temporal sequences into smaller, more manageable components for finer-grained feature learning.
For instance, Qiu et al. \cite{qiu2022spatio} proposed the Spatio-Temporal Tuple Transformer (STTFormer), which segments skeletal sequences into non-overlapping tuples, with each tuple representing a sub-action. Features are extracted from these sub-actions using a self-attention module and then aggregated through an inter-frame feature aggregation mechanism. Similarly, Do et al. \cite{do2025skateformer} partitioned joints and frames based on skeletal temporal relationships, distinguishing between adjacent and distant frames. This approach applies skeletal temporal self-attention within each region, effectively capturing both local and global temporal dynamics.

\subsection{Spatial Feature Extraction}
\textbf{Spatial Feature Extraction Using RNNs.}
Recurrent Neural Networks (RNNs) excel in temporal modeling but are relatively weaker in spatial feature extraction. To address this limitation, RNNs are often combined with spatial partitioning and hierarchical processing \cite{du2015hbrnn}, or integrated with other models such as CNNs, GCNs, and Transformers for auxiliary spatial modeling.
Liu et al. \cite{liu2016spatio} proposed a tree-structured skeleton traversal approach to enhance spatial information extraction, incorporating an LSTM with gates to effectively handle noise and occlusions. 
In their subsequent work \cite{liu2017global}, they introduced the Global Context-Aware Attention LSTM (GCA-LSTM), an attention-based network that leverages global context to selectively focus on informative joints. The GCA-LSTM architecture comprises two LSTM layers: the first encodes skeleton sequences to generate global context memory, while the second refines this memory using attention-based representations.
Zhang et al. \cite{zhang2017view} addressed viewpoint variations in skeleton sequences by integrating a temporal attention residual module within an LSTM framework. This module adjusts skeleton sequences to align viewpoints before the main LSTM network extracts further features.
Additionally, Li et al. \cite{li2021memory} combined attention-based RNNs with CNNs to enhance spatial modeling, demonstrating the potential of hybrid architectures in overcoming spatial modeling limitations.

\textbf{Spatial Feature Extraction Using CNNs.}
Convolutional Neural Networks (CNNs) excel in spatial feature extraction from 2D images but face challenges in skeleton-based HAR, where 3D skeleton data must be converted into pseudo images for processing.
Liu \cite{liu2017enhanced} utilized enhanced skeletal visualization to represent skeleton data. Wang et al. \cite{wang2018action} proposed the Joint Trajectory Map (JTM), which encodes the spatial structure and dynamics of joint trajectories into three texture images using color encoding, though this approach risks complexity and information loss. Li et al. \cite{li2019learning} introduced a shape-motion representation derived from geometric algebra, emphasizing the importance of joints and bones to maximize the information encoded in skeleton sequences.  Caetano et al. \cite{caetano2019skelemotion} developed SkeleMotion, a motion-based representation that explicitly encodes temporal dynamics by calculating the magnitude and direction of skeletal joint movements. 
Additionally, based on SkeleMotion, Caetano \cite{caetano2019skeleton} extended the framework by incorporating a tree structure and reference joints to represent skeleton images.
Despite these advancements, pseudo-image-based methods often face limitations. The convolution kernels typically learn co-occurrence features only from adjacent joints within their receptive field, potentially overlooking latent correlations between distant joints. To address this issue, partitioning or hierarchical feature extraction methods have been proposed to enhance learning, as demonstrated by  \cite{jia2020two,li2018co}.

\textbf{Spatial Feature Extraction Using Spatial Graphs.}
In recent years, due to the excellent expressive power of graph structures, graph-based learning models have garnered widespread attention \cite{zhao2019t, monti2018motifnet}. Researchers have focused on two main areas of improvement: better graph construction and more effective feature aggregation.

\begin{figure}[t]
	\centering
	\includegraphics[width=1.0\linewidth]{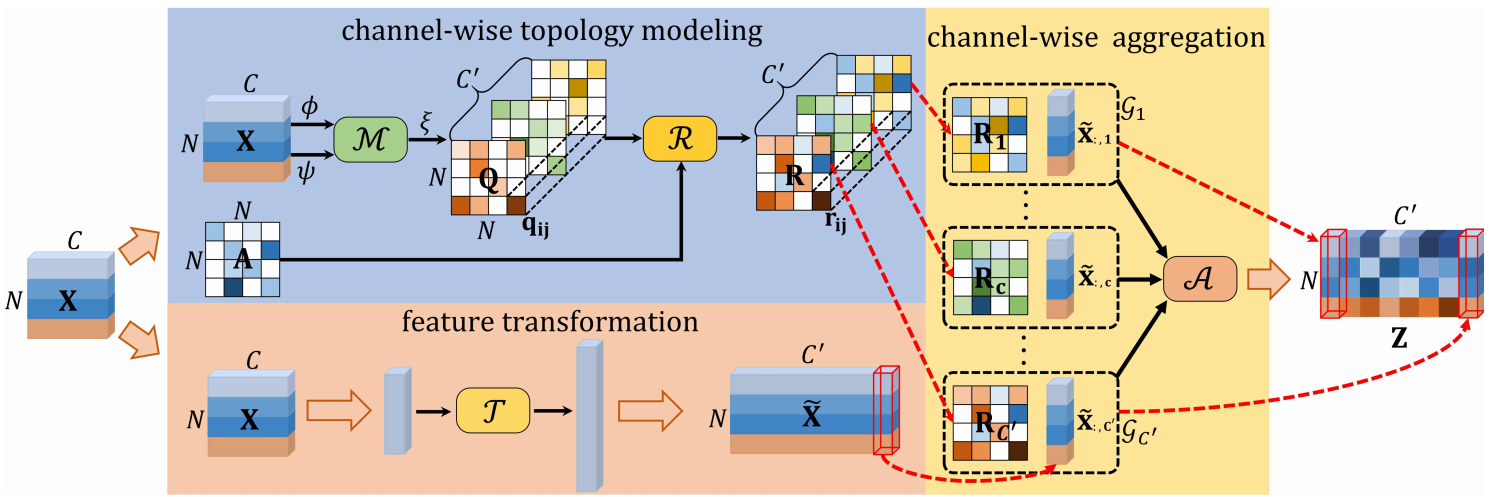}
	\caption{Architecture of the channel-wise topology refinement graph convolution.}
	\label{fig:ctrgcn}
	\vspace{-0.4cm}
\end{figure}

Improvements in graph construction typically rely on adjacency matrices that have been refined to capture not only physical connections but also more nuanced dynamic and semantic relationships.
Shi et al. \cite{shi2019two} introduced Two-stream Adaptive GCN (2sAGCN), learning graph topology via backpropagation and integrating joint coordinates with bone-based information to enhance action modeling. Li et al. \cite{li2019actional} proposed AS-GCN, combining action-specific and structural links into a generalized skeleton graph. Chen \cite{chen2021channel} developed CTR-GCN, grouping channels to share skeleton topology, learning shared topologies, and channel-specific correlations for efficient feature extraction, as shown in Fig. \ref{fig:ctrgcn}. Duan \cite{duan2022dg} presented DG-GCN, dynamically learning joint correlations with an affinity matrix, refining joint relations without predefined graphs, and enhancing precision through dot-product and normalization.


Improvements in feature aggregation focus on capturing both local and global spatial dependencies while enhancing extraction flexibility and efficiency. Wu et al. \cite{wu2019spatial} used cross-domain spatial residual layers and densely connected blocks to learn global features. Attention mechanisms were applied in \cite{si2019attention,wen2019graph} to extract discriminative information and capture global dependencies.
Traditional graph convolutions, limited by fixed receptive fields, were addressed by Shift-GCN \cite{cheng2020shiftgcn}, which expanded each node's receptive field to the entire skeleton graph via shift operations, improving flexibility. Chi et al. \cite{chi2022infogcn} introduced InfoGCN, incorporating an information bottleneck and attention-based graph convolution for dynamic, context-aware skeleton topologies. 
To capture long-term spatial dependencies, Peng et al. \cite{peng2021tripool} proposed the Tripool operation and loss function, enhancing efficiency. Liu et al. \cite{liu2023tsgcnext} advanced the field with TSGCNeXt, expanding skeleton feature channels fourfold to map features into a higher-dimensional space for more distinct spatial dependencies.

\textbf{Spatial Feature Extraction Using Attentions.}
By employing spatial self-attention modules, the model captures correlations between nodes within a frame and dynamic relationships across frames \cite{plizzari2021spatial, zhang2021stst}. Zhang et al. \cite{zhang2023skeletal} proposed RE-DMT, which captures global spatial features through dynamic masking and rearrangement. Liu et al. \cite{liu2023painet} designed PAINet, which learns discriminative features by combining intra-sequence and inter-sequence correlations, capturing co-motion and semantic relationships through a topology encoding module. 
The combination of Transformer and GCN is a key direction for spatial feature extraction. Shi \cite{shi2021star} used sparse multi-order topology matrix multiplication and self-attention to learn spatial relationships. Liu et al. \cite{liu2022graph} introduced the kernel attention adaptive graph Transformer network (KA-AGTN), which captures high-order dependencies between joints. Zheng et al. \cite{zheng2024spatio} proposed the spatio-temporal Dijkstra Attention (STDA) mechanism to enhance topological integration and optimize feature aggregation by strengthening local neighborhood connections.

\textbf{Spatial Feature Extraction Hierarchically.}
Similar to temporal hierarchical feature extraction, 
Spatial features can be extracted by partitioning the space into different receptive fields \cite{qiu2022spatio}, enabling the learning of spatial features at varying scales.
Li et al. \cite{li2018co} proposed a hierarchical method to learn co-occurrence features. It independently learned point-level features for each joint, which were then used as channels for convolution layers to learn hierarchical co-occurrence features. 
Liu et al. \cite{liu2020disentangling} integrated a disentangled multi-scale aggregation scheme with a spatio-temporal graph convolution operator called G3D, which served as a powerful feature extractor.
Wang \cite{wang20233mformer} formed a hypergraph to simulate hyperedges between graph nodes, introducing various pairings such as 'channel-temporal block', 'order-channel-body joint', and 'channel-hyper-edge (any order)', as well as 'channel-only' pairs to capture more complex spatial and temporal relationships.
Do et al. \cite{do2025skateformer} utilized the spatial relationships of the human skeleton to differentiate between proximal and distant joints. By combining this spatial distinction with temporal segmentation based on proximity, they defined four types of spatiotemporal relationships. This approach enables spatial self-attention within each region, effectively capturing both local and global temporal dependencies.

\textbf{Spatial Feature Extraction by Partition.}
Human motion is composed of independent or coordinated movements of various body parts, such as arms, legs, and the torso. For example, activities like running and swimming require precise coordination among these parts. To effectively recognize diverse human actions, it is essential to model the movements of individual body parts while capturing their collaborative dynamics. Methods such as \cite{wang2017modeling,du2015skecnn,li2017skeleton,ke2017skeletonnet,du2015hbrnn, shi2020decoupled}
decompose the skeleton into five parts (two arms, two legs, and the torso), modeling each part independently and then integrating their spatio-temporal features to extract collaborative motion patterns. Similarly, Avola et al. \cite{avola20192p} divides the body into upper and lower sections for separate processing.

However, this simple partitioning approach has limitations. It may overlook the overall common features of actions. To address this, Behera et al. \cite{behera2021co} proposed the Co-LSTM model, which uses a regularized deep LSTM network to learn co-occurrence features among skeletal joints, enhancing representational capacity.  
Additionally, part-level embeddings are often underutilized, especially for fine-grained actions. Wang et al. \cite{wang2023iip} proposed a Transformer-based network to model intra- and inter-part dependencies in the spatial domain for detailed feature representation. Chen et al. \cite{chen2022hierarchically} introduced a hierarchical Transformer for unsupervised learning, leveraging attention mechanisms on body part divisions.

\subsection{Assisted Feature Extraction}
Recently, numerous studies have explored applying large language models (LLMs) and generative models to skeleton-based action recognition to aid feature extraction. These approaches demonstrate vast potential, offering more innovative pathways for feature learning while significantly enhancing the robustness and adaptability of models.

\textbf{Assisted Feature Extraction Using LLMs.}
LLMs with their extensive knowledge base and powerful semantic reasoning, are increasingly vital in skeleton-based action recognition. They bridge the gap between visual features and high-level action semantics by providing semantic priors and generating multimodal information to enhance feature learning. For example, action descriptions generated by LLMs, such as body part movements, serve as supervisory signals in multimodal training, improving action feature representations and achieving state-of-the-art performance \cite{xiang2023generative}. Xu et al. \cite{xu2023language} combined LLM-generated semantic priors with multi-hop attention graph convolution networks to integrate global and category-level prior relationships, improving action representation quality.
LLMs can also act as auxiliary networks in multimodal frameworks, optimizing multimodal features during training while requiring only skeleton data during inference, balancing efficiency and performance \cite{liu2024multi}. Qu et al. \cite{qu2024llms} demonstrated that LLMs can directly recognize actions by converting skeleton sequences into action sentences, leveraging their language processing capabilities.

In tasks requiring recognition of unseen samples (e.g., zero-shot and few-shot), LLMs combine semantic reasoning with visual feature analysis to extract spatio-temporal patterns and semantic information. Zhou et al. \cite{zhou2023zero} used global alignment and temporal constraint modules to capture statistical correlations between visual and semantic spaces, enhancing recognition by incorporating temporal information. Yan et al. \cite{yan2024crossglg} proposed a global-local-global feature-guided strategy, generating high-level textual descriptions enriched with human knowledge to improve the recognition of unseen actions.

\textbf{Assisted Feature Extraction Using Generative Models.}
Generative models have also been widely applied to skeleton-based action recognition, particularly by modeling spatio-temporal dependencies through the prediction or reconstruction of masked skeleton data. For instance, Yan et al. \cite{yan2023skeletonmae} introduced the Skeleton Sequence Learning (SSL) framework, which uses the generative model SkeletonMAE to pre-train skeleton representations. This model reconstructs masked joints and edges by combining graph-based encoding with human body topology knowledge, effectively extracting skeleton features when combined with spatio-temporal learning modules. However, traditional generative models often face the challenge of redundant features during pre-training. To address this, Lin et al. \cite{lin2025idempotent} proposed the Idempotent Generative Model (IGM), which introduced idempotency constraints to enforce stronger consistency regularization in the feature space, resulting in more compact and effective feature representations.

Moreover, diffusion models, a type of generative model, have recently been adapted for skeleton-based action recognition. These models progressively map noise to the target distribution and have been modified for skeleton data with innovations such as specialized skeleton loss functions and the incorporation of Transformer architectures. Wu et al. \cite{wu2025macdiff} proposed the Masked Conditional Diffusion (MacDiff) framework, where the diffusion decoder conditions on representations from a semantic encoder. By applying random masking to the encoder's input, MacDiff reduces redundancy and introduced an information bottleneck, enhancing the quality and generalizability of skeleton representations.

\section{Spatial-temporal Modeling}
The most critical factors for skeleton-based action recognition lie in two key aspects: intra-frame representations that capture joint co-occurrences and inter-frame representations that model the temporal evolution of skeletons. Early research in this domain often focused on modeling either spatial or temporal features in isolation, neglecting the integration of both aspects. However, as the field has evolved, researchers have increasingly recognized the importance of jointly modeling spatial and temporal information. 

In recent years, numerous studies have aimed to capture the contextual dependencies in both the temporal and spatial domains, leveraging the spatial and temporal relationships among skeleton joints \cite{yussif2023srgcn,ijaz2022multi,chi2022infogcn}. 
This dual-focus approach has significantly advanced the development of the field. Consequently, current models typically incorporate architectural designs that integrate both spatial and temporal networks for skeleton-based action recognition.

\begin{figure}[t]
	\centering
	\includegraphics[width=0.95 \linewidth]{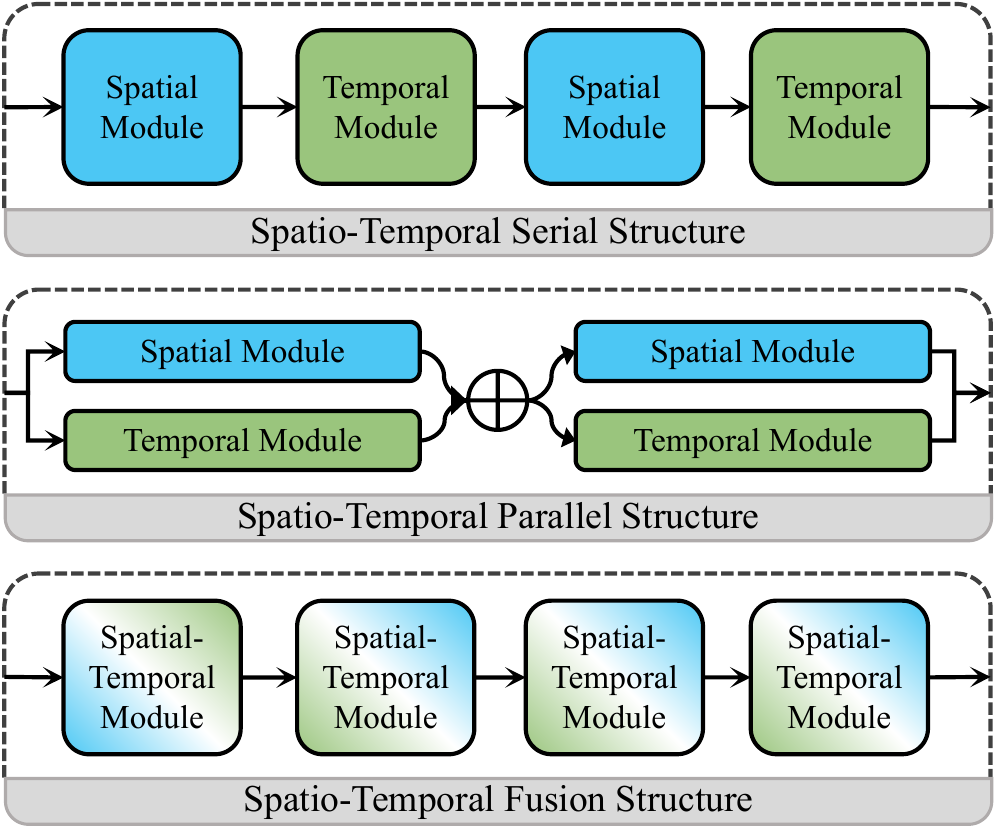}
	\caption{Schematic diagram of three types of spatial-temporal modeling architecture.}
	\label{fig:arch}
	\vspace{-0.4cm}
\end{figure}

Based on the arrangement of spatial and temporal modeling modules within their overall frameworks, these models can be broadly categorized into three types: \textbf{spatio-temporal serial structures}, \textbf{spatio-temporal parallel structures}, and \textbf{spatio-temporal fusion structures}. The schematic diagrams of these three architectures can be found in Fig. \ref{fig:arch}.

\subsection{Spatio-temporal Serial Structure}
\textbf{RNN-Based Methods.}
Du et al. \cite{du2015hbrnn} divided the human skeleton into five parts based on its physical structure, and then separately fed these parts into five distinct subnets. The representations extracted by these subnets are hierarchically fused to form the inputs for higher layers, enabling spatial modeling of the different body parts. The temporal dynamics of the entire body representation are then modeled using a bidirectional recurrent neural network (BRNN) \cite{schuster1997brnn}. Ultimately, these stacked BRNNs can be seen as extracting both spatial and temporal features from the skeleton sequences.
Song et al. \cite{song2017end} proposed an end-to-end multi-layer LSTM network incorporating both spatial and temporal attention mechanisms for action recognition. The network is designed to automatically select the most prominent joints within each frame using the spatial attention module, while the temporal attention module assigns varying levels of importance to different frames. Both modules are constructed using LSTM networks.
Zheng et al. \cite{zheng2019relational} proposed a two-stream ARRN-LSTM model, consisting of a Joints-stream and a Bone-stream. For each stream, they employed a Recurrent Relational Network (RRN) \cite{palm2018rrn} as a robust framework to learn spatial features from a single skeleton, followed by a multi-layer LSTM to capture the temporal dynamics within skeleton sequences. This approach enables the extraction of spatio-temporal features from both the Joints and Bone modalities.


\textbf{GCN-Based Methods.}
As one of the pioneering works, ST-GCN by Yan et al. \cite{yan2018stgcn} is the first to apply graph convolutional networks (GCNs) to skeleton-based action recognition. They construct a skeleton graph with spatial edges and temporal edges. The model then applies spatial graph convolution and temporal graph convolution sequentially to effectively model the dynamics of skeleton sequences.
Subsequent works have employed the same ST-GCN backbone \cite{shi2019two,cheng2020shiftgcn,zhang2020semantics,chen2021channel,yussif2023srgcn}, a classic paradigm for serial modeling along both the spatial and temporal dimensions.
Shi et al. \cite{shi2019two} proposed an adaptive graph convolutional network to more effectively learn the topology of the graph across different GCN layers.
Furthermore, Cheng et al. \cite{cheng2020shiftgcn} proposed two types of spatial shift graph operations for spatial graph modeling and two types of temporal shift graph operations for temporal modeling. This approach further reduces computational complexity while enhancing the receptive field for spatiotemporal modeling. 
Considering that semantic information, such as joint type and frame index, plays a crucial role in action recognition, Zhang et al. \cite{zhang2020semantics} proposed a Semantics-Guided Neural Network (SGN) that explicitly incorporates both semantics and dynamics  
for highly efficient action recognition.
Chen et al. \cite{chen2021channel} proposed a Channel-Wise Topology Refinement Graph Convolution (CTR-GC) that dynamically and effectively models channel-wise topology.
Yussif et al. \cite{yussif2023srgcn} proposed an instance-based self-relation modeling graph convolution that learns the connection strength between joints by determining their corresponding proximity in the feature space.

\textbf{Transformer-Based Methods.}
Observing that the temporal and spatial dimensions are fundamentally different, Shi et al. \cite{shi2020decoupled} introduced a novel decoupled spatial-temporal attention network (DSTA-Net) for skeleton-based action recognition. The skeleton sequence is first fed into a spatial attention module to model the spatial relationships between joints, and then sequentially passed through a temporal attention module to capture the temporal relationships between frames.
Sun et al. \cite{sun2021msst} proposed the MSST-RT framework, incorporating a lightweight relative transformer module for efficient spatial and temporal modeling. In the spatial dimension, the Spatial Relative Transformer (SRT) is designed to capture long-range dependencies while preserving the original skeleton topology. For the temporal dimension, the Temporal Relative Transformer (TRT) models long-range interactions between nonadjacent frames, ensuring that the sequential order of the skeleton remains intact. 
Ijaz et al. \cite{ijaz2022multi} designed a dual-modality transformer that leverages both acceleration and skeletal joint data. In each branch, a Spatial Transformer Encoder and a Temporal Transformer Encoder are connected sequentially to compute spatial and temporal features from the skeletal joints.
To simultaneously fuse human joint and part interactions, Wang et al. \cite{wang2023iip} proposed a novel spatial-temporal transformer network (IIP-Former), which efficiently captures both joint-level (intra-part) and part-level (inter-part) dependencies. By integrating spatial and temporal dependencies into a single transformer, IIP-Former reduces model complexity while enhancing generalization performance. 

\textbf{Hybrid-Network Based Methods.}
Recurrent Neural Networks (RNNs) demonstrate strong capabilities in capturing temporal dependencies but face limitations in effectively modeling the spatial coordination among different body parts. 
Hence, Si et al. \cite{si2018srtsl} proposed a Spatial Reasoning and Temporal Stack Learning (SR-TSL) model, which consists a a Spatial Reasoning Network (SRN) to extract high-level spatial structural information within each frame and a Temporal Stack Learning Network (TSLN) employing a composition of multiple skip-clip LSTMs to model fine-grained temporal dynamics across skeleton sequences, thus provieds a comprehensive solution for spatiotemporal representation learning.

Recognizing human actions effectively requires leveraging the intrinsic topology of joints, which provides critical contextual information. 
Chi et al. \cite{chi2022infogcn} proposed a novel approach that introduces a Self-Attention-based Graph Convolution (SA-GC) module. This module dynamically extracts the intrinsic graph structure while encoding a sequence of skeleton data. Following this, a Multi-Scale Temporal Convolution (MS-TC) module is employed to model temporal dynamics, enabling robust spatiotemporal representation learning for action recognition.
Liu et al. \cite{liu2022graph} introduced a Skeleton Graph Transformer (SGT) block featuring graph transformer operators designed to model higher-order spatial dependencies among joints. Compared to traditional feature extractors, the SGT block effectively alleviates the oversmoothing problem while capturing long-range dependencies with greater precision. Additionally, they propose a Temporal Kernel Attention (TKA) block, an efficient temporal feature enhancement module that enables the SGT blocks to focus on salient temporal features, thereby improving the overall spatiotemporal modeling capability.
Yang et al. \cite{yang2023stream} proposed the Stream-GCN network, which incorporates multi-stream components and channel attention to enhance spatial-temporal modeling. The multiple streams offer complementary representations by addressing different modalities, while the attention mechanism emphasizes the most critical channels. Each stream includes a novel cross-channel attention module for adaptively weighting channels, serving as the spatial modeling component, and integrates a multi-scale temporal modeling module to effectively capture temporal dynamics.


\subsection{Spatio-temporal Parallel Structure}

\textbf{RNN-Based Methods.} 
Wang et al. \cite{wang2017modeling} proposed a novel two-stream RNN architecture to model both temporal dynamics and spatial configurations for skeleton based action recognition. The temporal model, consisting of stacked LSTM for capturing spatial information of skeleton joints and hierarchical LSTM for processing temporal dynamics of individual body parts, integrates its outputs with those from the spatial model to produce the final action recognition results. 
Similarly, Cui et al. \cite{cui2018multi} proposed a multi-source model that integrates temporal and spatial models. The temporal model is further divided into three distinct branches to capture information at the global, local, and detailed levels, respectively, enabling a more comprehensive perception of action dynamics.

\textbf{CNN-Based Methods.}
Liu et al. \cite{liu2017two} are the first to apply 3D CNNs to skeleton-based action recognition. To enable 3D CNNs to learn robust features, skeleton joints are encoded into spatial and temporal volumes separately, capturing spatial and temporal information. Two independent 3D CNN models are trained individually for the spatial and temporal streams, which are fused only during the forward propagation phase.
Similarly, Xu et al. \cite{xu2018ensemble} designed four distinct sub-networks to extract diverse features, including global, local, focus, and motion features of actions. Among these, a Two-stream Entirety Net is proposed to capture both spatial and temporal features of the entire skeleton. In this architecture, the spatial stream extracts spatial features by sliding convolutional kernels in the spatial domain, while the temporal stream extracts temporal features by sliding the kernels in the temporal domain. The two streams are fused by computing cross-domain losses, enabling end-to-end training of this sub-network.

\textbf{GCN-Based Methods.}
Considering the intrinsic high-order correlations among skeleton joints is crucial for action recognition, Li et al. \cite{li2019spatio} proposed a novel spatio-temporal graph routing (STGR) scheme, consisting of two components: the Spatial Graph Router (SGR), which aims to discover the connectivity relationships among joints through sub-group clustering along the spatial dimension, and the Temporal Graph Router (TGR), which explores the structural information by measuring the correlation degrees between temporal 
trajectories. This framework effectively adapts to learn the intrinsic high-order connectivity relationships between physically distant skeleton joints.
Due to the complexity of human activities, coarse modelling features will lead to confusion between ambiguous actions with similar spatial appearances or temporal transformations. 
Hence, Zhou et al. \cite{zhou2023FRhead} proposed a Feature Refinement Head 
which first decouples the hidden feature maps into spatial and temporal components and then applies a contrastive learning loss with global class prototypes and ambiguous samples. Such spatial-temporal decouple module that mines the spatial and temporal information simultaneously to improve the discriminative ability of action representations.

\textbf{Transformer-Based Methods.} 
Zhang et al. \cite{zhang2021stst} introduced the Spatial-Temporal Specialized Transformer (STST), which independently models the temporal and spatial information of skeleton sequences while maintaining robustness to various abnormal scenarios. Specifically, this approach extracts coordinate information, semantic information, and temporal information into three distinct types of tokens to fully utilize the skeleton data without losing critical details. Building on a similar concept, Shi et al. \cite{shi2021star} proposed a novel skeleton-based action recognition model that applies sparse attention to the spatial dimension and segmented linear attention to the temporal dimension, further refining the efficiency and performance of attention mechanisms in this domain. 
Moreover, to align spatial and temporal information better, Liu et al. \cite{liu2023painet} proposed the Cross spatial alignment and the Cross temporal alignment branches to exploit inter-skeleton and intra-skeleton semantic information. Compared to previous works, their approach involves the alignment of the spatial and temporal domains within two parallel branches, enabling complementary attention to informative regions.

\textbf{Hybrid-Network Based Methods.}
Building upon the research foundation of ST-GCN, Plizzari et al. \cite{plizzari2021spatial} incorporated the concept of self-attention and proposed the ST-TR model, which incorporates a Spatial Self-Attention (SSA) module to capture intra-frame interactions between different body parts and a Temporal Self-Attention (TSA) module to model inter-frame correlations. The model employs a two-branch architecture, where each branch is trained independently, and the final scores are fused to generate predictions. 
To address the limitations of entangled spatio-temporal feature representation, Bai et al. \cite{bai2022hierarchical} proposed a novel architecture named Hierarchical Graph Convolutional Skeleton Transformer (HGCT). This model combines the strengths of Graph Convolutional Networks, including their ability to capture local topology, temporal dynamics, and hierarchical structure, with the global context modeling and dynamic attention capabilities of Transformers. By integrating these complementary features, HGCT facilitates the disentanglement of spatio-temporal representations, rather than solely focusing on refining graph topology design.

\textbf{Unsupervised Learning.} 
Zhou et al. \cite{zhou2023pstl} proposed a parallel spatiotemporal masking strategy to leverage local relationships within partial skeleton sequences, enhancing robustness to noise and data omissions. Dong et al. \cite{dong2023hico} introduced an instance-level representation, which combines temporal and spatial feature representations through concatenation, forming a unified and comprehensive representation to describe the entire skeleton sequence.
Since the complex spatiotemporal information extracted by most existing models is highly entangled, it becomes challenging to provide clear insights for subsequent analysis. To address this issue, Wu et al. \cite{wu2024scd} designed novel decoupling encoders that extract clean spatial and temporal representations from the otherwise entangled information, enabling more effective interpretation and comparison of the learned features.

\subsection{Spatio-temporal Fusion Structure}
\textbf{RNN-Based Methods.}
In addition to leveraging RNNs over the temporal domain to uncover discriminative dynamics and motion patterns for 3D action recognition, valuable discriminative information is also encoded in the static postures represented by the 3D locations of joints within individual frames. Furthermore, the sequential nature of skeleton data allows for the application of RNN-based learning in the spatial domain. Building on this observation, Liu et al. \cite{liu2016tree, liu2017tree} proposed a spatio-temporal LSTM (ST-LSTM) model. In the spatial domain, the body joints within a frame are processed sequentially, while in the temporal domain, the corresponding joints' locations are fed across time. Each unit in the model receives hidden representations from both preceding joints within the same frame and previous frames of the same joint, enabling the effective integration of spatial and temporal contextual information.
Despite its utility, LSTM exhibits limited attention capability for 3D action recognition. This limitation primarily arises from LSTM's inability to effectively perceive global contextual information, which is often crucial for addressing global classification tasks such as 3D action recognition. 
To overcome this constraint, Liu et al. \cite{liu2017global, liu2017skeleton} extended the original LSTM network by proposing the Global Context-Aware Attention LSTM (GCA-LSTM), a model designed to enhance attention capabilities for 3D action recognition. In the GCA-LSTM framework, global contextual information is introduced at every step of the network. This enables the model to evaluate the informativeness of new inputs at each step and adjust attention weights accordingly. 
This mechanism allows GCA-LSTM to focus more effectively on discriminative features critical to 3D action recognition.

\textbf{GCN-Based Methods.}
In previous work, a spatiotemporal layer typically combined spatial graph convolution and temporal convolution. Wu et al. \cite{wu2019sdgcn} introduced a cross-domain spatial residual layer as a residual branch to complement the original spatial graph convolution and temporal convolution branches.
Liu et al. \cite{liu2020msg3d} proposed G3D, a novel unified spatiotemporal graph convolution module that directly models cross-spacetime joint dependencies. They achieve this by introducing graph edges across the "3D" spatial-temporal domain as skip connections, enabling unobstructed information flow and significantly enhancing spatial-temporal feature learning.
Observing that many existing GCN methods use a pre-defined graph structure that remains fixed throughout the entire network, which can result in the loss of implicit joint correlations, Peng et al. \cite{peng2020learning} focused on reducing the manual effort required. They replace the fixed graph structure with a dynamic graph structure through automated neural architecture search (NAS) \cite{zoph2016neural}, and explore different graph generation mechanisms at various semantic levels. 
Inspired by the SlowFast network \cite{feichtenhofer2019slowfast} for RGB video recognition, Fang et al. \cite{fang2022spatial} proposed the Spatial-Temporal SlowFast Graph Convolutional Network (STSF-GCN). This model incorporates a fast pathway to capture short-range dependencies and a slow pathway to capture long-range dependencies, enabling the effective modeling of skeleton data within a unified spatiotemporal framework. Similar to MS-G3D \cite{liu2020msg3d} and GR-GCN \cite{gao2019optimized}, STSF-GCN leverages this dual-pathway approach to enhance its capability in capturing diverse spatiotemporal relationships.
Most previous works \cite{shi2019two,cheng2020decoupling,chen2021channel} treated all joints and edges as the same type, failing to capture the semantic properties of actions. To address this, Xie et al. \cite{xie2024dsgcn} proposed a dynamic semantic-based graph convolutional network (DS-GCN) for skeleton-based action recognition. In this approach, the joint and edge types are implicitly encoded in the skeleton topology. Specifically, two semantic modules are introduced: joint-type-aware adaptive topology and edge-type-aware adaptive topology. By integrating these semantic modules with temporal convolution, DS-GCN forms a powerful framework for action recognition.

\textbf{Transformer-Based Methods.}
Both recurrent and convolutional operations are local, neighborhood-based methods \cite{wang2018non}, operating in either space or time. These methods repeatedly extract and propagate local information to capture long-range dependencies. Despite the adoption of hierarchical networks \cite{du2015hbrnn,lee2017ensemble,cho2018spatio} to obtain deeper semantic representations, the challenge of capturing bidirectional semantic dependencies persists. To address this, Cho et al. \cite{cho2020self} introduced three novel self-attention networks (SAN), namely SAN-v1, SAN-v2, and SAN-v3, integrating Temporal Segment Networks (TSN) with these SAN variants. By capturing long-range correlations, their approach effectively extracts high-level semantics, allowing model to overcome the challenge of acquiring long-term semantic information.
Focusing solely on one- or few-hop graph neighborhoods may overlook dependencies between unconnected body joints. To address this, Wang et al. \cite{wang20233mformer} proposed using hypergraphs to model higher-order hyperedges between graph nodes (e.g., third- and fourth-order hyperedges capture interactions among three or four nodes), enabling the capture of more complex motion patterns involving groups of body joints.
Kim et al. \cite{kim2023cross} proposed a 3D deformable attention mechanism within a cross-modal learning framework, which addresses the spatio-temporal feature fusion challenge in skeleton-based action recognition. By leveraging an adaptive spatio-temporal receptive field and a cross-modal learning approach, their method effectively integrates spatial and temporal features across modalities.

Most transformer-based methods \cite{gao2022focal,xin2023mixformer,pang2022igformer,liu2023transkeleton} primarily focus on improving configuration and learning spatiotemporal correlations, without leveraging skeletal motion patterns in the frequency domain. This limitation hinders their ability to learn discriminative representations that capture subtle motion similarities. 
To address this issue, Wu et al. \cite{wu2024freq} proposed the Frequency-aware Mixed Transformer (FreqMixFormer), which is specifically designed to recognize similar skeletal actions with subtle, discriminative motions.
However, capturing correlations between all joints in all frames requires substantial memory resources.
Do et al. \cite{do2025skateformer} recently introduced the Skeletal-Temporal Transformer (SkateFormer), a novel approach that partitions joints and frames based on different types of skeletal-temporal relationships (Skate-Type). This method performs skeletal-temporal self-attention (Skate-MSA) within each partition, thereby reducing memory usage while maintaining effective learning of correlations.

\textbf{Hybrid-Network Based Methods.}
These methods typically integrate attention mechanisms to enable unified modeling of spatio-temporal features \cite{si2019attention,li2021memory,song2022constructing,xin2023mixformer,duan2023skeletr,zheng2024spatio,yun2024frfgcn}. 
Focus on how to effectively extract discriminative spatial and temporal features, Si et al. \cite{si2019attention} proposed a novel Attention Enhanced Graph Convolutional LSTM Network (AGCLSTM), consisting of a temporal hierarchical architecture to increase temporal receptive fields and an attention mechanism to enhance information of key joints.
Li et al. \cite{li2021memory} proposed the Temporal Attention Recalibration Module (TARM), which weights each frame and joint based on their importance within the action sequence. The Spatio-Temporal Convolution Module (STCM) is then used to model these enriched spatiotemporal features. This approach demonstrates the effectiveness of a novel "memory attention + convolution network" scheme for capturing the complex spatiotemporal variations in skeleton joints.
Observing that the recent State-Of-The-Art (SOTA) models for this task tends to be exceedingly sophisticated and over-parameterized, Song et al. \cite{song2022constructing} embedded recent advanced separable convolutional layers into an early fused Multiple Input Branches (MIB) network, constructing an efficient Graph Convolutional Network (GCN) baseline for skeleton-based action recognition.
Xin et al. \cite{xin2023mixformer} proposed the Spatial-Mix Module to dynamically capture multivariate topological relationships, and the Temporal-Mix Module, which employs multiple temporal models to ensure the richness of global differential expressions.
Duan et al. \cite{duan2023skeletr} targeted more general scenarios that typically involve a variable number of people and various forms of interaction between people and present SkeleTR, a unified solution for multiple skeleton-based action tasks.

\textbf{Mamba Based Methods.}
Mamba is an efficient state-space model (SSM) designed to handle sequential data and capture long-range dependencies. Martinel et al. \cite{martinel2024skelmamba} proposed SkelMamba, a skeleton-based action recognition model built on Mamba, which introduces the Time-Space Mamba Block (TSMB) to model temporal and spatial features in groups. The extraction of temporal features is accomplished through grouped 1D convolutions, incorporating time dynamics at various scales. SkelMamba extends the traditional Mamba's Selective Scan Mechanism (S6) by enabling four scanning directions: "time-to-space," "space-to-time," "reverse time-to-space," and "reverse space-to-time." This enhancement allows the model to outperform GCN and Transformer-based methods in capturing long-range dependencies, subtle motion features, and global dynamics. Additionally, the efficient SSM reduces computational complexity, making SkelMamba particularly suitable for skeleton action recognition, especially in medical applications that require detailed dynamic analysis.

\begin{table}[t]\scriptsize
	\centering
	\caption{The representative benchmark datasets with skeleton data modalities.}
 \scalebox{0.85}{
	\begin{tabular}{cccccc}
	\toprule
\textbf{Dataset}                                       & \textbf{Year} & \textbf{Class} & \textbf{Subject} & \textbf{Sample}  & \textbf{Viewpoint} \\
    \midrule
HDM0  \cite{muller2007documentation}                               & 2007  &130   & 5       & 2337    & 1 \\ 
MSR-Action3D  \cite{li2010action}                        & 2010  &20    & 10      & 567     & 1 \\ 
CAD-6  \cite{sung2011human}                              & 2011  &12    & 4       & 60      & - \\ 
MSR DailyActivity3D  \cite{wang2012mining}                  & 2012  &16    & 10      & 320     & 1 \\ 
UTKinect  \cite{xia2012view}                            & 2012  &10    & 10      & 200     & 1 \\ 
SBU Kinect Interactions  \cite{yun2012two}                & 2012  &8    & 7      & 282     & 1 \\
Berkeley MHAD  \cite{ofli2013berkeley}                       & 2013  &12    & 12      & 660     & 4 \\ 
CAD-120  \cite{koppula2013learning}                             & 2013  &10    & 4       & 120     & - \\ 
IAS-lab  \cite{munaro20133d}                             & 2013  &15    & 12      & 540     & 1 \\ 
J-HMDB  \cite{jhuang2013towards}                              & 2013  &21    & -       & 31838   & - \\ 
MSRAction-Pair  \cite{oreifej2013hon4d}                      & 2013  &12    & 10      & 360     & 1 \\ 
UCFKinect  \cite{ellis2013exploring}                           & 2013  &16    & 16      & 1280    & 1 \\ 
Multi-View TJU  \cite{liu2014multiple}                      & 2014  &20    & 22      & 7040    & 2 \\ 
Northwestern-UCLA  \cite{wang2014cross}                   & 2014  &10    & 10      & 1475    & 3 \\ 
UPCV  \cite{theodorakopoulos2014pose}                                & 2014  &10    & 20      & 400     & 1 \\ 
UWA3D Multiview  \cite{rahmani2014action}                     & 2014  &30    & 10      & 900     & 4 \\ 
SYSU 3D HOI  \cite{hu2015jointly}                         & 2014  &12    & 40      & 480     & 1 \\ 
TJU  \cite{liu2015coupled}                                 & 2015  &15    & 20      & 1200    & 1 \\ 
UTD-MHAD  \cite{chen2015utd}                           & 2015  &27    & 8       & 861     & 1 \\ 
UWA3D Multiview II  \cite{rahmani2016histogram}                  & 2015  &30    & 10      & 1075    & 4 \\ 
Large Scale Combined(LSC) \cite{zhang2018large}                     & 2016  & 88    & -      & 3898     & - \\
NTU RGB+D  \cite{shahroudy2016ntu}                           & 2016  &60    & 40      & 56880     & 80 \\
PKU-MMD  \cite{liu2017pku}                             & 2017  &51    & 66      & 1076    & 3 \\ 
Kinetics400  \cite{carreira2017quo}                    & 2017  &400    & -      & 300000    & - \\
RGB-D Varying-view  \cite{ji2018large}                           & 2018  &40    & 118      & 25600       &  8+1(360$^\circ$) \\ 
DHP19  \cite{calabrese2019dhp19}                               & 2019  &33    & 17      & -       & 4 \\ 
Drive\&Act  \cite{martin2019drive}                           & 2019  &83    & 15      & -       & 6 \\ 
MMAct  \cite{kong2019mmact}                               & 2019  &37    & 20      & 36764   & 4+Egocentric \\ 
NTU RGB+D 120  \cite{liu2019ntu}                      & 2019  &120   & 106     & 114480  & 155 \\ 
ETRI-Activity3D  \cite{jang2020etri}                     & 2020  &55    & 100     & 112620  & - \\ 
EV-Action  \cite{wang2020ev}                           & 2020  &20    & 70      & 7000    & 9 \\ 
IKEA ASM  \cite{ben2021ikea}                          & 2020  &33    & 48      & 16764   & 3 \\ 
UAV-Human  \cite{li2021uav}                          & 2021  &155   & 119     & 67428   & - \\ 
    \bottomrule
	\end{tabular}}
 \label{table_datasets}
\end{table}

\begin{table}[t]
\centering
\setlength\tabcolsep{3pt} 
	\caption{Performance of skeleton-based deep learning HAR methods on Kinetics400 dataset.}
 \scalebox{0.95}{
	\begin{tabular}{c|ccc}
	\toprule
\multicolumn{2}{c}{\textbf{Methods}}                          & Year & Accuracy  \\
    \midrule
\multirow{18}{*}{\rotatebox{90}{GCN-Based Methods}}
&AR-GCN  \cite{ding2019attention}                              & 2019  &33.5  \\ 
&ST-GR  \cite{li2019spatio}                                    & 2019  &33.6   \\ 
&PR-GCN  \cite{li2021pose}                                     & 2020  &33.7   \\ 
&PeGCN  \cite{yoon2022predictively}                            & 2020  &34.8   \\ 
&AS-GCN  \cite{li2019actional}                                 & 2019  &34.8   \\ 
&SLnL-rFA  \cite{hu2019skeleton}                               & 2019  &36.6  \\ 
&DGNN  \cite{shi2019skeleton}                                  & 2019  &36.9   \\ 
&JB-AAGCN  \cite{shi2020skeleton}                              & 2019  &37.4   \\ 
& GCN-NAS  \cite{peng2020learning}                             & 2020  &37.1   \\ 
& CGCN  \cite{yang2020unifying}                                & 2020  &37.5  \\\ 
& MS-AAGCN ~ \cite{shi2020skeleton}                            & 2020  &37.8   \\  
& Dynamic GCN  \cite{ye2020dynamic}                            & 2020  &37.9    \\ 
& MS-G3D  \cite{liu2020disentangling}                          & 2020  &38.0      \\ 
& DualHead-Net  \cite{chen2021learning}                        & 2021  &38.4      \\
& 2s-AGCN+TEM  \cite{obinata2021temporal}                      & 2021  &38.6     \\
& PoseConv3D  \cite{duan2022revisiting}                        & 2022  &49.1  \\ 
& DS-GCN \cite{xie2024dsgcn}                                   & 2024  &50.6  \\ 
& ProtoGCN \cite{liu2024revealing}                             & 2024  &51.9  \\ 
\midrule
\multirow{6}{*}{\rotatebox{90}{\scriptsize Other Msethods}}
&Ours-Conv-Chiral  \cite{yeh2019chirality}                     & 2019  &30.9  \\ 
&SLnL-rFA  \cite{hu2019skeleton}                               & 2019  &36.6  \\ 
& ST-TR-agcn  \cite{plizzari2021skeleton}                      & 2020  &37.4     \\ 
& STF \cite{ke2022towards}                                     & 2022  &39.9     \\ 
& PYSKL \cite{duan2022pyskl}                                   & 2022  &49.1     \\ 
&Structured Keypoint Pooling  \cite{hachiuma2023unified}       & 2023  &52.3     \\ 
    \bottomrule
	\end{tabular}}
 \label{tab:Kinetics400}
\end{table}

\begin{table}[!htbp]\scriptsize
\setlength\tabcolsep{3pt} 
\caption{Performance of skeleton-based deep learning HAR methods on NTU RGB+D and NTU RGB+D 120 datasets. ‘CS’, ‘CV’, and ‘CP’ denote Cross-Subject, Cross-View, and Cross-Setup evaluation criteria.}
\scalebox{0.8}{
\begin{tabular}{c|cccccc}
\toprule
\multicolumn{2}{c}{\multirow{4}{*}{\textbf{\small{Methods}}}}  & \multirow{4}{*}{\textbf{\small{Year}}}     & \multicolumn{4}{c}{\textbf{Dataset}}           \\   
\cmidrule{4-7}
\multicolumn{2}{c}{~}     &~        & \multicolumn{2}{c}{\textbf{NTU RGB+D 60}}     & \multicolumn{2}{c}{\textbf{NTU RGB+D 120}}      \\      
\cmidrule(lr{0pt}){4-5}
\cmidrule(lr{0pt}){6-7}
\multicolumn{2}{c}{~}     &~        & \textbf{CS} & \textbf{CV}                  & \textbf{CS}   & \textbf{CP} \\ 
\midrule
\multirow{16}{*}{\rotatebox{90}{RNN-Based Methods}}   &  HBRNN-L  \cite{du2015hierarchical}    & 2015        & 59.1        & 64.0        & -             & -      \\
                         &      P-LSTM  
                         \cite{shahroudy2016ntu}                    & 2016        & 62.9        & 70.3        & 25.5          & 26.3      \\
                         &      Trust Gate ST-LSTM 
                         \cite{liu2017skeleton}                  & 2016        & 69.2        & 77.7        & 58.2          & 60.9     \\
                         &  LSTM with Geometric Features  \cite{zhang2017geometric} & 2017        & 70.3        & 82.4      & -             & -      \\
                         &       Two-stream RNN  \cite{wang2017modeling}            & 2017        & 71.3        & 79.5        & -             & -      \\
                         &       STA-LSTM  \cite{song2017end}         & 2017        & 73.4        & 81.2        & -             & -      \\
                         &       GCA-LSTM  \cite{liu2017global}       & 2017        & 74.4        & 82.8        & 58.3          & 59.2     \\
                         &   Ensemble TS-LSTM  \cite{lee2017ensemble} & 2017        & 74.6           & 81.3      & -             & -      \\
                         &  Bayesian GC-LSTM  \cite{zhao2019bayesian} & 2018        & 81.8        & 89.0        & -             & -      \\
                         &       MANs  \cite{li2021memory}            & 2018        & 82.7        & 93.2        & -             & -      \\
                         &       IndRNN  \cite{li2018independently}   & 2018        & 86.7        & 93.7        & -             & -      \\
                         &       VA-RNN (aug.)  \cite{zhang2017view}  & 2019        & 79.8        & 88.9        & -             & -      \\
                         &       ARRN-LSTM  \cite{zheng2019relational}& 2019        & 81.8        & 89.6        & -          & -     \\
                         &       AGC-LSTM  \cite{si2019attention}     & 2019        & 89.2        & 95.0        & -             & -      \\
                         &       Logsig-RNN  \cite{liao2021logsig}    & 2021        & -           & -           & 68.3          & 67.2     \\
                         &       KShapeNet  \cite{friji2021geometric} & 2021        & 84.2        & 89.7        & 74.8          & 76.9     \\
                         \midrule
\multirow{19}{*}{\rotatebox{90}{CNN-Based Methods}}  & JTM  \cite{wang2016action}   & 2016  & 73.4     & 75.6     & -          & -      \\
                         &       Res-TCNs  \cite{kim2017interpretable}& 2017        & 74.3        & 83.1        & 67.5          & 75.6      \\
                         &       SkeletonNet  \cite{ke2017skeletonnet}& 2017        & 75.9        & 81.2        & -             & -      \\
                         &       JDMs  \cite{li2017joint}             & 2017        & 76.2        & 82.3        & -             & -      \\
                         &       Clips+CNN+MTLN  \cite{ke2017new}     & 2017        & 79.6        & 84.8        & 58.4          & 57.9   \\
                         &  Enhanced Skeleton Visualization  \cite{liu2017enhanced} & 2017        & 80.0        & 87.2      & 60.3      & 63.2     \\
                         &Translation-Scale Invariant Mapping  \cite{li2017skeleton}& 2017        & 85.0        & 92.3      & -         & -      \\
                         &       RotClips+MTCNN  \cite{ke2018learning}& 2018        & 81.1        & 87.4        & 62.2          & 61.8      \\
                         &       Ensem-NN  \cite{xu2018ensemble}      & 2018        & 84.8        & 91.2        & -             & -      \\
                         &       HCN  \cite{li2018co}                 & 2018        & 86.5        & 91.1        & -             & -      \\
                         &       TSRJI  \cite{caetano2019skeleton}    & 2019        & 73.3        & 80.3        & 67.9          & 62.8      \\
                         & SkeleMotion  \cite{caetano2019skelemotion} & 2019        & 76.5        & 84.7        & 67.7          & 66.9     \\
                         &       Skepxels  \cite{liu2019skepxels}     & 2019        & 81.3        & 89.2        & -             & -      \\
                         &       Shape-motion  \cite{li2019learning}  & 2019        & 82.9        & 90.0        & -             & -      \\
                         &       VA-CNN (aug.)  \cite{zhang2017view}  & 2019        & 88.7        & 94.3        & -             & -      \\
                         &       Gimme Signals  \cite{memmesheimer2020gimme} & 2020 & -           & -           & 71.9          & 83.5      \\
                         &       TS-TCNs  \cite{jia2020two}           & 2020        & 82.4        & 90.2        & 71.9          & 83.5      \\
                         &  Fuzzy Integral-Based CNN  \cite{banerjee2020fuzzy} & 2020     & 84.2     & 89.7     & 74.8          & 76.9      \\
                         &       SEMN  \cite{wang2021skeleton}        & 2021        & 82.0        & 85.8        & -             & -      \\
                         \midrule
\multirow{25}{*}{\rotatebox{90}{GCN-Based Methods}}     &   ST-GCN  \cite{yan2018stgcn}  & 2018  & 81.5   & 88.3  & -     & -   \\
                         &       SR-TSL  \cite{si2018srtsl}        & 2018        & 84.8        & 92.4        & -             & -      \\
                         &       AS-GCN  \cite{li2019actional}        & 2019        & 86.8        & 94.2        & -             & -      \\
                         &       2s-AGCN  \cite{shi2019two}           & 2019        & 88.5        & 95.1        & -             & -      \\
                         &       DGNN  \cite{shi2019skeleton}         & 2019        & 89.9        & 96.1        & -             & -      \\
                         &     DC-GCN+ADG  \cite{cheng2020decoupling} & 2020        & 88.2        & 95.2        & 90.8          & 96.6    \\
                         &       SGN  \cite{zhang2020semantics}       & 2020        & 89.0        & 94.5        & 79.2          & 81.5    \\
                         &       GCN-NAS  \cite{peng2020learning}     & 2020        & 89.4        & 95.7        & -             & -      \\
                         &    4s Shift-GCN  \cite{cheng2020shiftgcn}  & 2020        & 90.7        & 96.5        & 85.9          & 87.6      \\
                         &    MS-G3D  \cite{liu2020disentangling}     & 2020        & 91.5        & 96.2        & 86.9          & 88.4      \\
                         &       Sym-GNN  \cite{li2021symbiotic}      & 2021        & 90.1        & 96.4        & -             & -      \\
                         &       Else-Net  \cite{li2021else}          & 2021        & 91.6        & 96.4        & -             & -      \\
                         &       CTR-GCN  \cite{chen2021channel}      & 2021        & 92.4        & 96.8        & 88.9          & 90.      \\
                         &       SATD-GCN  \cite{zhang2022spatial}    & 2022        & 89.3        & 95.5        & -             & -      \\
                         & EfficientGCN-B4  \cite{song2022constructing} & 2022      & 92.1        & 96.1        & 88.7          & 88.9      \\
                         &       TD-GCN  \cite{liu2023temporal}       & 2022        & 92.8        & 96.8        & 89.8          & 91.2      \\
                         &       TCA-GCN  \cite{wang2022skeleton}     & 2022        & 92.8        & 97.0        & 89.4          & 90.8      \\
                         &   PSUMNet  \cite{trivedi2023psumnet}       & 2022        & 92.9        & 96.7        & 89.4          & 90.6      \\
                         &Language Supervised Training  \cite{xiang2022language} & 2022        & 92.9        & 97.0     & 89.9          & 91.1      \\       
                         &      InfoGCN  \cite{chi2022infogcn}        & 2022        & 93.0        & 97.1        & 89.8          & 91.2      \\
                         &         DG-STGCN  \cite{duan2022dg}        & 2022        & 93.2        & 97.5        & 89.6          & 91.3      \\
                         & HD-GCN  \cite{lee2022hierarchically}       & 2022        & 93.4        & 97.2        & 90.1          & 91.6      \\
                         &   PoseC3D  \cite{duan2022revisiting}       & 2022        & 94.1        & 97.1        & -             & -      \\
                         &      TSGCNeXt  \cite{liu2023tsgcnext}      & 2023        & 93.1        & 97.0        & 90.2          & 91.7      \\
                         &       LA-GCN  \cite{xu2023language}        & 2023        & 93.5        & 97.2        & 90.7          & 91.8      \\
                          &       BlockGCN  \cite{zhou2024blockgcn}        & 2024        & 93.1        & 97.0       & 90.3          & 91.5      \\    
                         &       DeGCN  \cite{myung2024degcn}        & 2024        & 93.6        & 97.4        & 91.0          & 92.1      \\
                     
                         \midrule
\multirow{12}{*}{\rotatebox{90}{Transformer-Based Methods}}& TS-SAN  \cite{cho2020self} & 2020  & 87.2    & 92.7       &  -     &  - \\
                         &       DSTA-Net  \cite{shi2020decoupled}    & 2020        & 91.5        & 96.4      & 86.6            & 89.0      \\
                         &   Sparse Transformer  \cite{shi2021star}   & 2021        & 83.4        & 84.2      & 78.3            & 78.5      \\
                         &       ST-TR  \cite{plizzari2021spatial}    & 2021        & 89.9        & 96.1      & 81.9            & 84.1      \\
                         &       STST  \cite{zhang2021stst}           & 2021        & 91.9        & 96.8      & -               & -      \\
                         &     IIP-Transformer  \cite{wang2021iip}    & 2021        & 92.3        & 96.4      & 88.4            & 89.7     \\
                         &       KA-AGTN  \cite{liu2022graph}    & 2022        & 90.4        & 96.1      & 86.1            & 88.0      \\
                         &   STTFormer  \cite{qiu2022spatio}          & 2022        & 92.3        & 96.5      & 88.3            & 89.2      \\
                         &    Hi-TRS  \cite{chen2022hierarchically}   & 2022        & 90.0        & 95.7      & 85.3            & 87.4      \\
                         &   Hyperformer  \cite{zhou2022hypergraph}   & 2022        & 92.9        & 96.5      & 89.9            & 91.3      \\
                         &    STAR-Transformer  \cite{ahn2023star}    & 2023        & 92.0        & 96.5      & 90.3            & 92.7      \\
                         &    TemPose  \cite{ibh2023tempose}          & 2023        & 92.7        & 95.2      & 88.5            & 87.0      \\
                         &    SkateFormer  \cite{do2025skateformer}          & 2024        & 93.5        & 97.8      & 89.8            & 91.4      \\
                         &     FreqMixFormer \cite{wu2024freq}          & 2024        & 93.6        & 97.4      & 90.5            & 91.9      \\                         
\bottomrule
\end{tabular}}
\label{tab: ntu}
\vspace{-1em}
\end{table}

Over the past decade, the field has accumulated many classic datasets, summarized in Table \ref{table_datasets}. Skeleton sequence datasets such as MSR Actions 3D \cite{li2010action}, 3D Action Pairs \cite{oreifej2013hon4d}, and MSR Daily Activity 3D \cite{wang2012mining} have been extensively analyzed in numerous previous surveys. In addition to these well-known and widely used datasets, this survey primarily focuses on the current mainstream and challenging datasets.

\textbf{SYSU3D Human-Object Interaction Set (SYSU)} \cite{hu2015jointly}. The Kinect-captured dataset consists of 12 actions performed by 40 subjects, with a total of 480 sequences. Each subject is represented by 20 joints. This dataset presents a challenge due to the high similarity among the activities, making it difficult to distinguish between them based solely on skeletal data.

\textbf{UWA3D Multi-view Activity II (UWA3D)} \cite{rahmani2016histogram}. This Kinect-captured dataset contains 30 actions performed by 10 different subjects. The videos are captured from 4 different views: front view (V1), left side view (V2), right side view (V3), and top view (V4). It has 1075 sequences in total. This dataset is challenging because of the diversity of viewpoints, self-occlusion, and high similarity among activities. 

\textbf{Northwestern-UCLA (N-UCLA)} \cite{wang2014cross}.This Kinect-captured dataset contains 1494 videos of 10 actions. These actions are performed by 10 subjects and repeated 1 to 6 times. There are three views. Each subject has 20 joints. 

\textbf{SBU Kinect Interaction (SBU)} \cite{yun2012two}. This Kinect-captured dataset is an interaction dataset with each action performed by two subjects. The dataset contains 282 sequences across 8 action classes, with each subject having 15 joints. It's important to note that both the SYSU and SBU datasets are captured using a single camera from one primary viewpoint. However, actions are performed by different subjects at varying locations, distances from the camera, and orientations, which introduces additional variability in the data.

\textbf{SHREC’17 Track}  \cite{de2017shrec}. The dataset contains 2,800 gesture sequences, performed in two variations: using one finger and the whole hand. Each gesture is repeated between 1 and 10 times by 28 participants, with the hand skeleton consisting of 22 joints.
The dataset follows the same evaluation protocol as in \cite{baek2016kinematic, wang2017modeling}: training data consists of 1,960 sequences, while the remaining 840 sequences are used for testing. The gesture sequences can be labeled into 14 or 28 classes, depending on the number of fingers used and the gesture represented.

\textbf{DHG-14/28} \cite{de2016skeleton}. The DHG-14/28 dataset is collected using the Intel Real-Sense camera and contains 2,800 sequences of 14 gestures, performed 5 times by 20 participants. Evaluation follows a leave-one-subject cross-validation strategy.

\textbf{Human3.6M}  \cite{ionescu2013human3}. It has 15 types of actions performed by 7 actors (S1, S5, S6, S7, S8, S9, and S11). Each pose has 32 joints in the format of an exponential map. We convert them to 3D coordinates and angle representations and discard 10 redundant joints. The global rotations and translations of poses are excluded. The frame rate is downsampled from 50fps to 25fps. S5 and S11 are used for testing and validation respectively, while the remaining are used for training.

\textbf{The Kinetics-400} \cite{carreira2017quo}. The dataset is a large-scale video dataset collected from YouTube videos with 400 action classes. It contains 250K training and 19K validation 10-second video clips.
The performance of all deep learning methods on the Kinetics-400 dataset is summarized as shown in Table \ref{tab:Kinetics400}. 

\textbf{NTU-RGB+D} \cite{shahroudy2016ntu}.
Proposed in 2016, the dataset contains 56,880 video samples collected using Microsoft Kinect v2, making it one of the largest datasets for skeleton-based action recognition. It provides 3D spatial coordinates for 25 joints in each human action. Two evaluation protocols are recommended: Cross-Subject and Cross-View. In the Cross-Subject protocol, which includes 40,320 samples for training and 16,560 for evaluation, the 40 subjects are split into training and evaluation groups. In the Cross-View protocol, it includes 37,920 samples for training and 18,960 for evaluation, using cameras 2 and 3 for training and camera 1 for evaluation.


\textbf{NTU-RGB+D 120} \cite{liu2019ntu}.
Recently, an extended version of the original NTU-RGB+D dataset, known as NTU-RGB+D 120, has been introduced. This dataset comprises 120 action classes and 114,480 skeleton sequences, with an expanded range of 155 viewpoints. Specifically, CS refers to Cross-Subject, CV denotes Cross-View in NTU-RGB+D, and CSet (Cross-Setting) is used in NTU-RGB+D 120. Table \ref{tab: ntu} summarizes the performance of various deep learning methods on both NTU-RGB+D and NTU-RGB+D 120 datasets. While existing algorithms have achieved impressive results on the original NTU-RGB+D dataset, NTU-RGB+D 120 remains a significant challenge. Current skeleton-based action recognition methods perform relatively poorly on this dataset, making it one of the most demanding benchmarks in the field.


\textbf{UCF101} \cite{soomro2012ucf101}. 
The UCF101 dataset is a relatively small benchmark dataset, building on the UCF50 dataset. It consists of 13,320 video clips categorized into 101 action classes, which are further grouped into five main types: Body Motion, Human-Human Interactions, Human-Object Interactions, Playing Musical Instruments, and Sports. The total duration of the video clips exceeds 27 hours. All videos were sourced from YouTube, featuring a consistent frame rate of 25 FPS and a resolution of 320 × 240.

\textbf{HMDB51} \cite{kuehne2011hmdb}.
This dataset contains 6,766 video clips across 51 action categories, sourced from movies and web content. Each category includes at least 101 clips, making it a diverse collection of realistic videos.
The dataset follows an original evaluation protocol involving three distinct training/testing splits. In each split, 70 clips per action category are designated for training, and 30 clips for testing.

\section{Future Work}
In the previous sections, we review the various deep learning methods and datasets for HAR with skeleton data modality. Next, we will discuss some potential and important research directions, which need further development and exploration.

\textbf{Hand gesture recognition.} The joints in the human hand are numerous and very dense. Hand gestures are clear signs that express individual action intentions and emotions well. It also has a strong auxiliary role in recognizing other non-gesture actions. Although GCN-based methods \cite{liu2023temporal}  \cite{song2022dynamic} achieved high success, two problems in gesture recognition need to be solved. (1) Most existing databases condense the hand skeleton into 1 or 2 skeleton points, which is far from accurate gesture recognition. Although several datasets dedicated to gesture recognition have been proposed, they tend to contain only gestures, which is too narrow in the application scope. (2) If the joints of the hand were fully incorporated into the human skeleton topology, the existing methods would incur considerable additional computational overhead and redundant information, because most of the time the joints of the hand are less active than those of the rest of the human body.

\textbf{Few-shot/one-shot action recognition.}
In natural situations, people perform a wide range of actions at varying frequencies. Some actions may be rare and difficult to capture, yet they can be highly significant. For instance, a "robbery" is an infrequent event in daily life, but when it occurs in video surveillance, it must be recognized immediately.
One-shot recognition aims to find a method to classify new instances with a single reference sample. Although there have been some attempts \cite{wang2022temporal} \cite{ma2022learning}, they still cannot solve the problem of data scarcity in certain practical scenarios.
Possible approaches for solving problems of this category are metric learning \cite{hoffer2015deep,wang2014learning}, or meta-learning  \cite{finn2017model}. 
In action recognition, this suggests that a novel action can be learned from a single reference demonstration. However, unlike one-shot image classification, skeleton-based actions involve sequential data, where a single frame often lacks sufficient context to recognize a new activity. Current solutions \cite{memmesheimer2022skeleton} primarily focus on static images, overlooking the importance of image sequences. Static images are inadequate for distinguishing between similar actions, such as sitting down and standing up, as they fail to capture the temporal dynamics that differentiate these actions.
Except for accurately distinguishing between categories of actions when data is scarce, the difficulty with one-shot action recognition is to distinguish between unknown classes that have never been seen before. Berti  \cite{berti2022one} adds a discriminator to determine the unknown action class. Besides, video-based occluded one-shot recognition  \cite{peng2023delving} is still attractive to be researched.

\textbf{Human Motion Prediction}
Human actions exhibit continuity and causality. Motion prediction aims to observe partially completed behaviors to infer subsequent actions, which is crucial for preventing potentially risky behaviors.
Earlier methods \cite{tang2018long}  \cite{sang2020human} mainly used RNN-based approaches, which were difficult to train due to discontinuity and error accumulation problems. The current mainstream approach is based on GCN \cite{ma2022progressively} and Transformer \cite{mao2022weakly, guo2023back}. It is common practice to copy the last observable frame action multiple times (the length of the predicting action) to convert the original input sequence into an extended input sequence.
Then the ground truth of future poses is also appended to the observable poses to obtain the extended ground truth of predicting action. Experiments show that the prediction between the extended sequences is easier than between the original sequences and the former makes the model's prediction more accurate. Although existing methods are capable of predicting a single action, they need to be improved for predicting a long time complex action. The key point is how to simulate the smooth transition between two continuous actions.

\textbf{Real-time action recognition.}
The existing HAR methods achieve excellent results based on many model parameters and computational overhead. In real application scenarios, such as embedded devices, devices have to quickly obtain real-time human action information.
Therefore how to improve the recognition efficiency of HAR methods and reduce resource consumption is eager for further research. 

\textbf{Two-person interaction recognition and group actions recognition.}
Most  existing datasets record actions for a single person, and few studies \cite{pang2022igformer} \cite{perez2021interaction} \cite{wen2023interactive} \cite{wen2024chase} focus on the recognition of two-person interaction recognition. The reality is that people gather together, and most actions captured by the camera are group actions.  For two-person interaction or group actions, the urgent problem is how to capture the information of individual interaction. While conventional approaches focus on modeling temporal and spatial information within individuals, modeling information between individuals requires new skeleton topologies or additional neural networks.

\textbf{Unsupervised Learning.}
Manually annotating data is tedious work, so labeled data is expensive. However, deep learning methods usually require large amounts of data for training models Lacking the amount of data leads to over-fitting problems and worse performance.
Meanwhile, unsupervised learning techniques can train models with unlabeled data which greatly reduces the need for labeled data.
Due to the relatively easy collection of unlabeled data, unsupervised learning techniques have become a significant research direction  that is worthy of exploring in the future.

Earlier unsupervised learning methods for skeleton-based action recognition can be divided into two categories: using RNN-based encoder-decoders and contrastive learning schemes. Several existing methods utilize RNN-based encoder-decoder networks  \cite{zheng2018unsupervised}  \cite{su2020predict}. The decoder of these networks performs a pre-training task to induce the encoder to extract an appropriate representation for action recognition. RNN-based models suffer from long-range dependencies, and the GCN-based models have a similar challenge because they deliver information sequentially along a fixed path  \cite{plizzari2021spatial}. Therefore, the RNN and GCN-based methods have limitations in extracting global representations from the motion sequence, especially from long motions. The transformer is more popular because of its ability to model the local dynamics of joints and capture the global context from motion sequences. How to use pre-training strategies more efficiently to generate motion sequence representations that match the natural structure of the human body remains to be studied.
Other methods exploit the contrastive learning scheme  \cite{rao2021augmented}  \cite{wang2022contrast}  \cite{lin2023actionlet}. These methods augment the original motion sequence and regard it as a positive sample while considering other motion sequences as negative samples. The model is then trained to generate similar representations between the positive samples using contrastive loss. How to construct and match positive and negative samples and calculate their similarity is critical.
\section{Conclusions}

In this paper, we presented a comprehensive review of skeleton-based action recognition, focusing on the core challenges of transforming unstructured data into structured representations and modeling the spatial features of skeletal sequences. Moving beyond traditional model architecture classifications, we adopted a task-oriented framework that emphasizes key stages such as data representation, joint modeling, and spatiotemporal feature modeling. This approach provides a more holistic view of the task, offering valuable insights into the intrinsic challenges and the research opportunities they present.
We highlighted the crucial role of skeleton data preprocessing, including derived modalities and data augmentation, which directly influence the success of subsequent spatiotemporal modeling. Additionally, we explored emerging cutting-edge methods such as hybrid architectures, Mamba models, and large language models (LLMs), offering a fresh perspective on the future directions of research. By refining the task into progressively detailed sub-tasks, our review encourages researchers to address common challenges from a data-driven perspective, fostering innovation in model design and optimization.
Ultimately, this review aims to deepen the understanding of skeleton-based action recognition and to guide future advancements in the field, especially as it continues to evolve with new technologies and methodologies.

\small
\bibliographystyle{IEEEtran}
\bibliography{ref}
\end{document}